\documentclass[sigconf]{acmart}

\AtBeginDocument{%
  }

\copyrightyear{2022}
\acmYear{2022}
\setcopyright{acmlicensed}\acmConference[SA '22 Conference
Papers]{SIGGRAPH Asia 2022 Conference Papers}{December 6--9, 2022}{Daegu,
Republic of Korea}
\acmBooktitle{SIGGRAPH Asia 2022 Conference Papers (SA '22 Conference
Papers), December 6--9, 2022, Daegu, Republic of Korea}
\acmPrice{15.00}
\acmDOI{10.1145/3550469.3555383}
\acmISBN{978-1-4503-9470-3/22/12}

\usepackage{graphicx}
\usepackage{amsmath} 
\usepackage{color}

\usepackage{booktabs}
\usepackage{multirow}
\usepackage{xcolor}
\usepackage{soul}
\usepackage{pifont}
\usepackage{hyperref}
\usepackage{wrapfig}
\usepackage{threeparttable}

\usepackage{amsmath,amsfonts,bm}

\def\eqref#1{equation~\ref{#1}}

\def\1{\bm{1}}

\def\rvc{{\mathbf{c}}}

\def\rvr{{\mathbf{r}}}

\def\vc{{\bm{c}}}
\def\vd{{\bm{d}}}

\def\vo{{\bm{o}}}

\def\vr{{\bm{r}}}

\def\vt{{\bm{t}}}

\def\vv{{\bm{v}}}

\def\mV{{\bm{V}}}

\DeclareMathAlphabet{\mathsfit}{\encodingdefault}{\sfdefault}{m}{sl}
\SetMathAlphabet{\mathsfit}{bold}{\encodingdefault}{\sfdefault}{bx}{n}

\def\sR{{\mathbb{R}}}

\def\usecolor{0}
\if\usecolor1
    \definecolor{revcolor}{rgb}{0,0,1}
\else
    \definecolor{revcolor}{rgb}{0,0,0}
\fi
\newcommand{\rev}[1]{#1}
\newcommand{\revb}[1]{\textcolor{revcolor}{#1}}
\definecolor{amber}{rgb}{1.0, 0.75, 0.0}

\makeatletter
\DeclareRobustCommand\onedot{\futurelet\@let@token\@onedot}
\def\@onedot{\ifx\@let@token.\else.\null\fi\xspace}

\def\eg{\emph{e.g}\onedot} 
\def\ie{\emph{i.e}\onedot} 
 
\def\etc{\emph{etc}\onedot}

\makeatother

\usepackage{xspace}
\def\name{TiNeuVox\@\xspace}
\begin{document}

\title{Fast Dynamic Radiance Fields with Time-Aware Neural Voxels}

\author{Jiemin Fang}
\authornote{Equal contributions.}
\orcid{0000-0002-0322-4582}
\affiliation{%
  \institution{Institute of AI \& School of EIC, HUST}
  \city{Wuhan}
  \country{China}
  }
\email{jaminfong@hust.edu.cn}

\author{Taoran Yi}
\authornotemark[1]
\orcid{0000-0001-7316-7547}
\affiliation{%
  \institution{School of EIC, HUST}
  \city{Wuhan}
   \country{China}
  }
\email{taoranyi@hust.edu.cn}

\author{Xinggang Wang}
\authornote{Corresponding author.}
\orcid{0000-0001-6732-7823}
\affiliation{%
  \institution{School of EIC, HUST}
  \city{Wuhan}
   \country{China}
  }
\email{xgwang@hust.edu.cn}

\author{Lingxi Xie}
\orcid{0000-0003-4831-9451}
\affiliation{%
  \institution{Huawei Inc.}
  \city{Beijing}
   \country{China}
  }
\email{198808xc@gmail.com}

\author{Xiaopeng Zhang}
\orcid{0000-0001-6337-5748}
\affiliation{%
  \institution{Huawei Inc.}
  \city{Shanghai}
   \country{China}
  }
\email{zxphistory@gmail.com}

\author{Wenyu Liu}
\orcid{0000-0002-4582-7488}
\affiliation{%
  \institution{School of EIC, HUST}
  \city{Wuhan}
   \country{China}
  }
\email{liuwy@hust.edu.cn}

\author{Matthias Nie{\ss}ner}
\orcid{0000-0001-6093-5199}
\affiliation{%
  \institution{Technical University of Munich}
  \city{Munich}
   \country{Germany}
  }
\email{niessner@tum.de}

\author{Qi Tian}
\orcid{0000-0002-7252-5047}
\affiliation{%
  \institution{Huawei Inc.}
  \city{Shenzhen}
   \country{China}
  }
\email{tian.qi1@huawei.com}

\renewcommand\shortauthors{Fang, Yi, et al.}

\ccsdesc[500]{Computing methodologies~3D imaging}



\begin{teaserfigure}
\vspace{-10pt}
\centering
  \includegraphics[width=0.92\textwidth]{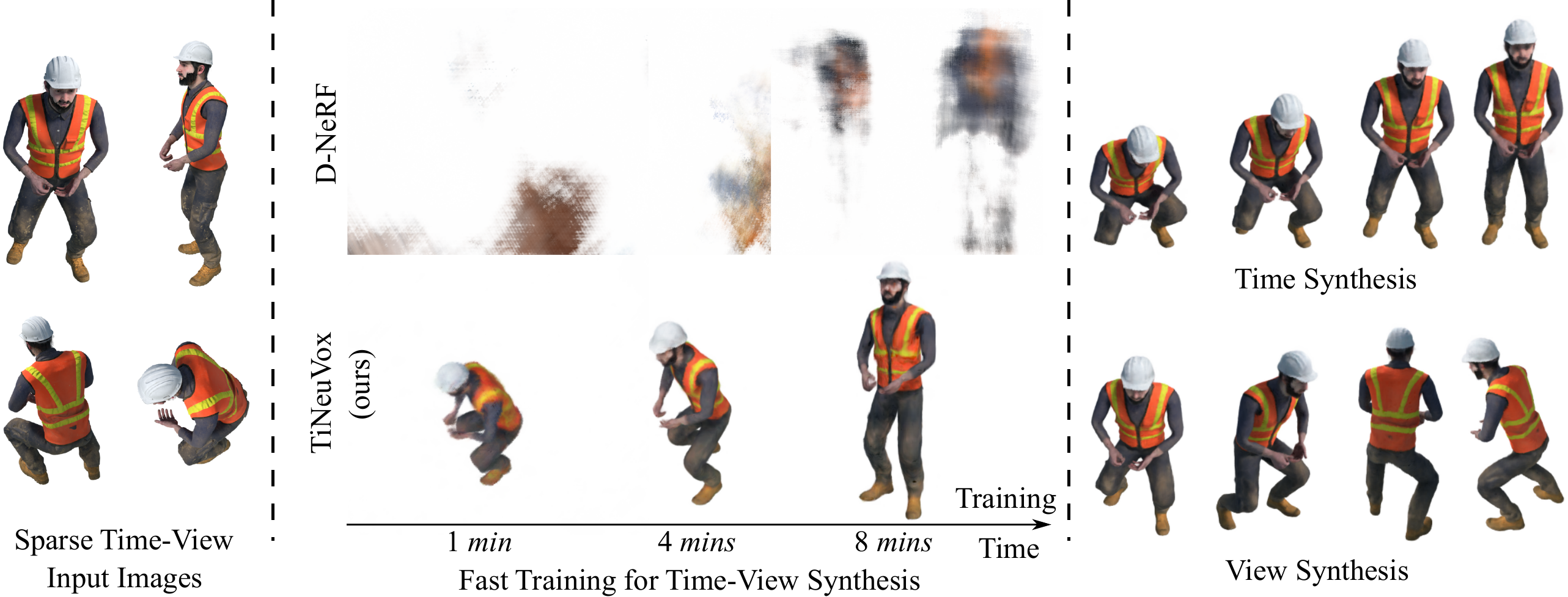}
  \vspace{-10pt}
  \caption{We propose a radiance field framework equipped with time-aware neural voxels, which can learn dynamic scenes with an extremely fast convergence speed. Comparisons with D-NeRF~\cite{pumarola2021d} are shown. Sparse time-view images are taken and novel time and view images can be synthesized with our method.}
  \Description{We propose a radiance field framework equipped with time-aware neural voxels, which can learn dynamic scenes with an extremely fast convergence speed.}
  \label{fig: teaser}
\end{teaserfigure}

\begin{abstract}
Neural radiance fields (NeRF) have shown great success in modeling 3D scenes and synthesizing novel-view images. However, most previous NeRF methods take much time to optimize one single scene. Explicit data structures, \eg voxel features, show great potential to accelerate the training process. However, voxel features face two big challenges to be applied to dynamic scenes, \ie modeling temporal information and capturing different scales of point motions. We propose a radiance field framework by representing scenes with time-aware voxel features, named as \name. A tiny coordinate deformation network is introduced to model coarse motion trajectories and temporal information is further enhanced in the radiance network. A multi-distance interpolation method is proposed and applied on voxel features to model both small and large motions. Our framework significantly accelerates the optimization of dynamic radiance fields while maintaining high rendering quality. Empirical evaluation is performed on both synthetic and real scenes. Our \name completes training with only 8 minutes and 8-MB storage cost while showing similar or even better rendering performance than previous dynamic NeRF methods. Code is available at \url{https://jaminfong.cn/tineuvox}.
\end{abstract}

\maketitle

\vspace{-7pt}
\section{Introduction}
Rendering plays a critically important role in various applications, \eg virtual reality, interactive gaming, and movie production \etc. High-quality and fast rendering techniques bring users realistic experience and make more applications possible. Recent neural rendering methods, represented by NeRF (neural radiance fields)~\cite{mildenhall2020nerf}, have shown great power for modeling 3D scenes with compact implicit representations and synthesizing high-quality novel-view images. However, conventional NeRF methods~\cite{mildenhall2020nerf,Barron_2021_ICCV} mainly focus on static scenes, while real-life scenarios usually involve object motions or topological changes. A series of subsequent NeRF works~\cite{pumarola2021d,park2021nerfies,li2021nsff,Tretschk_2021_ICCV,park2021hypernerf} improve radiance field construction towards dynamic scenes. 

Besides, fast training and rendering speed of NeRF is needed in real-life applications. Conventional NeRF methods bear large time and computation cost to optimize the field networks, \ie dozens of hours in general. Especially, most existing methods model dynamic scenes by introducing an additional deformation network with a similar scale of the radiance network, which maps point coordinates into a canonical space. This manner means much more cost for training and inferring dynamic fields. The cumbersome time cost impedes wide applications in real-life scenarios.

Representing scenes with explicit data structures shows great success in dramatically accelerating NeRF training and rendering \cite{Hedman_2021_ICCV,sun2021direct,yu_and_fridovichkeil2021plenoxels,muller2022instant}. However, it is challenging to represent dynamic scenes with explicit structures from two main aspects. On the one hand, these scenes involve complicated point motions where encoding temporal information is required. One direct and simple solution is to expand the voxel grids with an additional time dimension. However, this manner will undoubtedly increase memory cost significantly. \rev{Changing voxel grids from 4D to 5D, \ie $(C, N_x, N_y, N_z) \to (C, N_x, N_y, N_z, N_t)$, will multiply the storage cost by $N_t$}. On the other hand, there usually exist motions of different scales. Voxels with high resolutions locate in small grids, which fail to model large motions; voxels in large grids fail to capture details with small motions.

To tackle the above challenges, we propose a new dynamic radiance field method, named as \name, by representing scenes with time-aware voxel features. To encode temporal information, we first build a highly compressed deformation network which maps 3D point coordinates into a coarse canonical space. Voxel features are queried with the transformed coordinates. We further enhance the temporal information by feeding time and coordinate embeddings into the latter radiance network. Thus deviation introduced by point mapping can be automatically suppressed by the neural network. Moreover, we propose a multi-distance interpolation method, where features are obtained from voxels with multiple distances. In this way, both small and large motions can be modeled even though only one single-resolution voxel features are constructed.

We summarize our contributions as follows.
\vspace{-4pt}
\begin{itemize}
    \item We are the first to represent dynamic scenes with optimizable explicit data structures, which shows extremely high training efficiency.
    \item We encode coarse point motions with a tiny deformation network and enhance temporal information in the radiance network. A multi-distance interpolation technique is proposed to model both small and large motions with one single resolution of voxel features.
    \item We evaluate our method on both synthetic and real scenes, where our \name achieves better or similar rendering quality with $8$-MB storage by taking only 8 minutes, $150\times$ faster than D-NeRF~\cite{pumarola2021d} and $192\times$ faster than Hyper-NeRF~\cite{park2021hypernerf}.
\end{itemize}

\vspace{-7pt}
\section{Related Works}
\subsection{Neural Rendering for Dynamic Scenes}
The emergence of NeRF (neural radiance field)~\cite{mildenhall2020nerf} has greatly boosted the development of rendering techniques. A series of subsequent works improve NeRF from various aspects, \eg anti-aliasing~\cite{Barron_2021_ICCV}, camera parameter optimization~\cite{wang2021nerf,lin2021barf}, rendering large-scale unbounded scenes~\cite{kaizhang2020}, and reconstructions from unstructured image collections~\cite{Martin-Brualla_2021_CVPR} \etc.

Building radiance fields on dynamic scenes is one of the most important branch of improved NeRF which are tightly related to real-world scenarios. The key problem to solve dynamic-scene rendering lies in temporal information encoding. One stream of dynamic NeRF methods~\cite{li2021nsff,xian2020space,gao2021dynamic,du2021neural} model deformations in scenes by extending radiance fields with an additional time dimension. Due to prior knowledge lacking for structures in non-rigid scenes, additional geometry regularization and data modalities need to be introduced. Another class of methods~\cite{pumarola2021d,park2021nerfies,park2021hypernerf} introduce an additional deformation field to predict motions of points by mapping point coordinates into a canonical space, where large motions or geometry changes can be captured and learned.
Other methods improve dynamic neural rendering from various aspects, including distinguishing fore- and back-ground~\cite{tretschk2021non}, strengthening quality via depth information~\cite{attal2021torf} and producing sharper results by setting up key frames~\cite{li2021neural} \etc. \rev{A series of articulated NeRF methods~\cite{su2021nerf,weng2022humannerf,xu2021h,noguchi2021neural} are also proposed to represent human body motions.} Most of current dynamic NeRF methods still bear cumbersome training cost. We dramatically accelerate the training speed by introducing optimizable explicit voxel features, while a compressed deformation network along with temporal information enhancement is designed. The overall framework achieves a good quality-speed trade-off via proper computation allocation on explicit and implicit representations.

\begin{figure*}[thbp]
    \centering
    \includegraphics[width=0.9\linewidth]{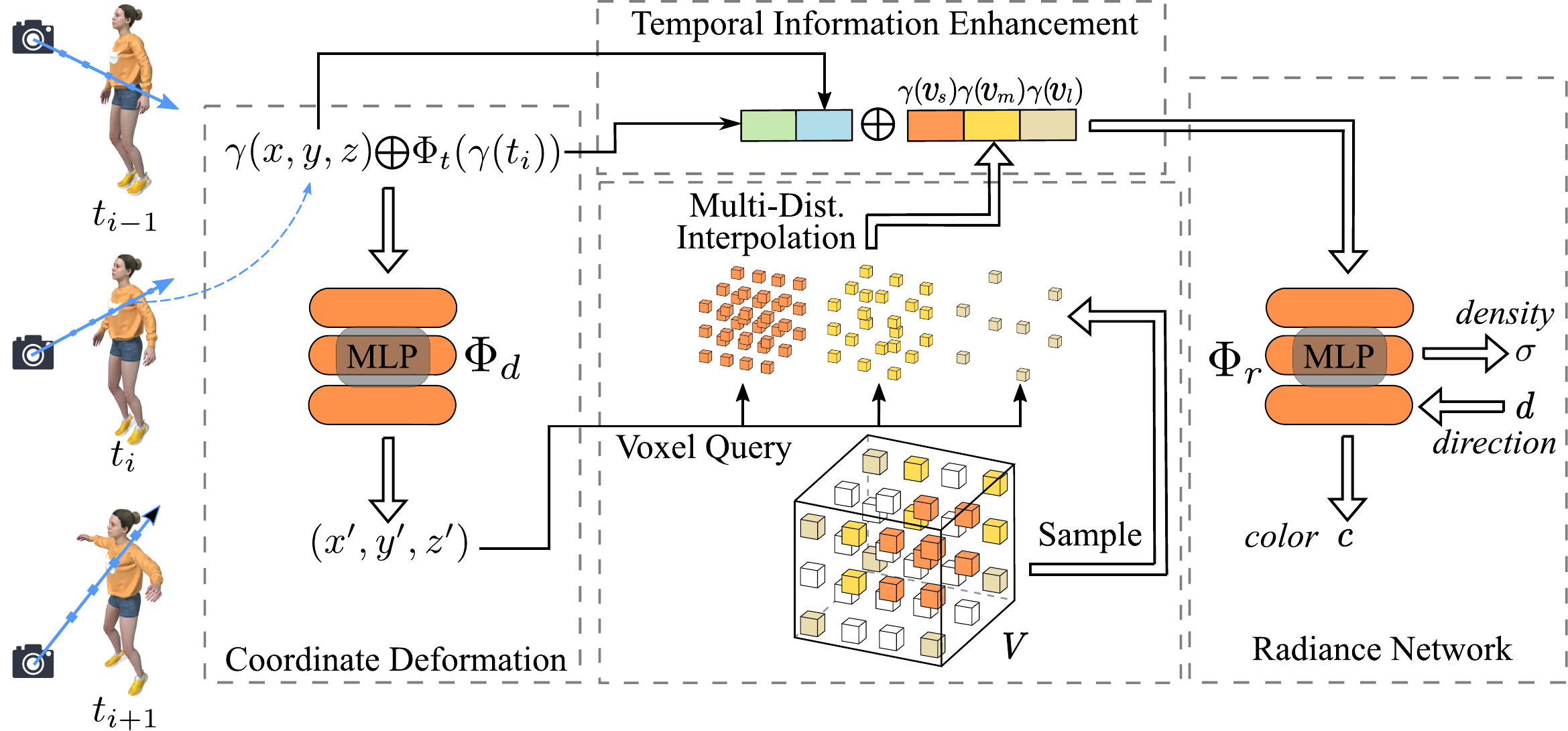}
    \vspace{-10pt}
    \caption{Overall framework of \name. First, a deformation network $\Phi_d$ takes both point coordinates $\gamma(x, y, z)$ and encoded time embeddings $\vt_i = \Phi_t(\gamma(t_i)$ as input to obtain the shifted coordinates $(x', y', z')$. Then voxel features in grids with different sampling strides are queried and interpolated according to deformed coordinates. To enhance temporal information, coordinates $\gamma(x, y, z)$ and time embeddings $\vt_i$ are further concatenated with interpolated voxel features, \rev{$\gamma(\vv_s)$, $\gamma(\vv_m)$, and $\gamma(\vv_l)$}, which are finally fed into the radiance network to produce the density $\sigma$ and color $\vc$.}
    \vspace{-10pt}
    \Description{Overall framework of TiNeuVox.}
    \label{fig: framework}
\end{figure*}

\vspace{-5pt}
\subsection{Neural Rendering Acceleration}
\paragraph{Rendering Acceleration}
Though conventional NeRF methods show high rendering quality, it bears high latency as a series of points along each ray need to be sampled and inferred for volume rendering. Some works propose to reduce inference times for acceleration by improving sampling strategies \cite{neff2021donerf,lindell2021autoint,piala2021terminerf,arandjelovic2021nerf,fang2021neusample} or introducing efficient rendering techniques \cite{sitzmann2021lfns}. \rev{\citet{LiuGLCT20,Hedman_2021_ICCV,Yu_2021_ICCV,Garbin_2021_ICCV,Reiser_2021_ICCV,wizadwongsa2021nex,sitzmann2019deepvoxels}} store properties like densities produced by pre-trained radiance fields into explicit data structures, \eg voxel grids or MPIs (multiplane images). Only a few points need to be inferred by a small network for view-dependent color predictions. Though these methods have achieved real-time rendering performance, they still bear immense pre-training cost and cumbersome additional storage cost.

\vspace{-10pt}
\paragraph{Convergence Acceleration}
Some methods explore to reduce training cost from the generalization perspective. \citet{Chen_2021_ICCV,yu2021pixelnerf,wang2021ibrnet,liu2022neuray,wang2022fourier} substantially pre-train NeRF on various scenes to obtain generalizable properties or features. \cite{deng2021depth} achieves faster training speed with external depth information. Some works~\cite{sun2021direct,yu_and_fridovichkeil2021plenoxels} propose to represent scenes with explicit voxel-grid features/properties and directly optimize these voxels for extremely fast convergence speed, reducing training time from hours to minutes. However, storage cost is significantly increased for storing voxel features. Recent works effectively reduce the storage cost via voxel hashing~\cite{niessner2013real,muller2022instant}, tensor decomposition~\cite{kolda2009tensor,chen2022tensorf} and bitrate dictionary lookup~\cite{vbnerf} while still maintaining surprisingly high training speed. These voxel-optimizable methods yet only focus on static scenes, where voxel features are direct to construct for only spatial information. We introduce optimizable explicit voxel features into dynamic scenes. Temporal information is encoded to obtain time-aware neural voxel features. Our \name achieves similar or better rendering performance than previous dynamic NeRF methods with training time reduced from days to $8$ minutes.

\vspace{-5pt}
\section{Method}
In this section, we first review methodologies of the original NeRF \cite{mildenhall2020nerf} in Sec~\ref{subsec: prel}. Second, we describe how we represent dynamic scenes with explicit voxel features in Sec.~\ref{subsec: neuvox}. Then we propose to encode temporal information along with voxel features in Sec.~\ref{subsec: temp_enc}. Finally, the overall framework and optimization procedures are presented in Sec.~\ref{subsec: frame_optim}.

\vspace{-5pt}
\subsection{Preliminaries}
\label{subsec: prel}
Neural radiance fields are first proposed in \cite{mildenhall2020nerf}, which models 3D scenes by mapping the coordinate $(x,y,z)$ and view direction $(\theta, \phi)$ of each point in the space into its color $\vc$ and density $\sigma$. The mapping function is usually instantiated as a neural network $\Phi_r$. This process can be formulated as
\vspace{-5pt}
\begin{equation}
    \vc,\sigma = \Phi_{r}{(x,y,z,\theta, \phi)}.
\end{equation}

To get the expected color $C(\vr)$ of the pixel in the image captured by the camera, a ray $\vr(t)=\vo+t\vd$ marching from the center of the camera to the pixel is involved, where $\vo$ and $\vd$ are the origin and direction of the ray respectively. $t$ denotes the distance from one point to the camera, which ranges from a pre-defined near bound $t_{n}$ to far bound $t_{f}$. The pixel color is rendered by sampling a series of points along the ray and performing the classical volume rendering~\cite{kajiya1984ray}:
\vspace{-5pt}
\begin{equation}
\begin{aligned}
\label{eq: rendering}
    \hat{C}(\vr) &= \sum_{i=1}^N T_i(1 - \text{exp}(-\sigma_i \delta_i))\vc_i,\\
    T_i &= \text{exp}(- \sum_{j=1}^{i-1} \sigma_j  \delta_j),
\end{aligned}
\end{equation} 
where $\delta_i$ is the distance between the $i_\text{th}$ and $(i+1)_\text{th}$ sample point, $N$ denotes the number of sampled points. Eq.~\ref{eq: rendering} connects real 3D points with image pixels by accumulating colors $\vc_i$ and densities $\sigma_i$ of sample points along the ray. Finally, radiance fields are optimized via gradient descent by minimizing the following loss:
\begin{equation}
    \label{eq: color_loss}
    \mathcal{L} = \lVert\hat{C}(\vr) - C(\vr)\rVert^2_2 ,
\end{equation}
where $C(\vr)$ denotes the groundtruth color of the pixel.

Besides, NeRF~\cite{mildenhall2020nerf} finds that details cannot be depicted by merely inputting $x,y,z$ coordinates and $\theta, \phi$ view directions. A positional encoding is introduced to map the input into a periodic formulation.
\begin{equation}
    \label{eq: pe}
    \gamma(x) =(sin(2^0x),cos(2^0x),...,sin(2^{L-1}x),cos(2^{L-1}x)),
\end{equation}
where $L$ is a hyperparameter that controls the highest frequency of the input.

\vspace{-5pt}
\subsection{Multi-Distance Interpolation with Neural Voxels}
\label{subsec: neuvox}
Conventional NeRF methods \cite{mildenhall2020nerf,Barron_2021_ICCV,pumarola2021d,park2021nerfies,park2021hypernerf} build radiance fields with pure implicit representations, \ie neural networks. Though this manner achieves promising rendering quality with high storage efficiency, it usually takes non-negligible time cost to optimize the fields, \eg dozens of hours or even several days. To accelerate the convergence of radiance fields, we propose to represent scenes with explicit data structures except for implicit ones. To this end, \textbf{neural voxels} are introduced, which are a set of features organized as morphology of voxel grids. These features are designed to be further queried and inferred by neural networks to obtain certain properties, such as radiance and transmittance. As shown in Fig.~\ref{fig: framework}, a scene is represented with grids of neural voxels $\mV \in \sR^{C_v \times N_x \times N_y \times N_z}$, where $C_v$ denotes the channel number of each voxel feature, and $N_x$, $N_y$ and $N_z$ denote the length of three spatial dimensions. 

\begin{wrapfigure}{l}{0.5\linewidth}
    \vspace{-10pt}
    \centering
    \includegraphics[width=1\linewidth]{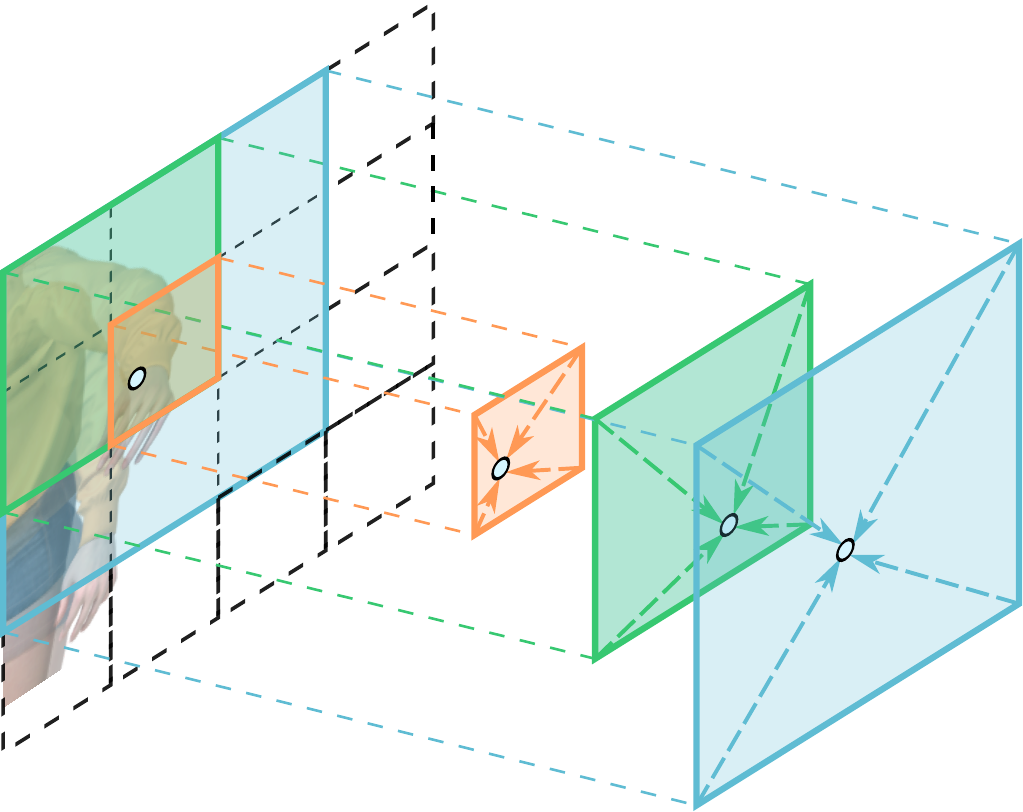}
    \vspace{-20pt}
    \caption{Illustration of multi-distance interpolation.
    }
    \Description{Illustration of multi-distance interpolation. Voxel features are interpolated from grids sampled with different strides.}
    \vspace{-10pt}
    \label{fig: interp}
\end{wrapfigure}

To predict the property of one 3D point, neural voxels stored in eight vertices of the grid this point lies in are queried and interpolated trilinearly. The interpolated feature is then inferred by neural networks to predict the expected properties. Considering points in dynamic scenes may move with a dramatic motion trajectory, small grids of neural voxels have limited capacity to model these point movements. We propose a \textbf{multi-distance interpolation} method to model point motions with various scales. As shown in Fig.~\ref{fig: interp}, besides the smallest grid, voxel features are also interpolated from vertices of larger grids. This means the final features for inference not only come from nearest voxels, but also from sub- and subsub-nearest voxels. In this way, small motions can be modeled via near voxels while motions in a large region are perceived with farther voxels. 

We implement the above process as follows. When performing interpolation, we sample neural voxels from the pre-built grids $\mV$ with different strides $s_1, s_2, s_3, \dots$. Then features are trilinearly interpolated with several sampled voxel grids respectively. Finally, these features are concatenated and fed into neural networks. This process can be formulated as
\rev{\begin{equation}
\begin{aligned}
\label{eq: interp}
    \vv   &= \vv_1 \oplus \cdots \vv_m \cdots \oplus \vv_M,\\
    \vv_m &= \text{interp} (x, y, z, \mV[::s_m]),
\end{aligned}
\end{equation}}
where $M$ denotes the total number of defined sampling strides, ${x, y, z}$ denote the coordinates of the 3D point.

Moreover, before being fed into the neural network for inference, these interpolated voxel features are first positional-encoded as in Eq.~\ref{eq: pe}. This manner is of critical importance for compressing neural voxels into small sizes while maintaining strong performance for modeling details, which is evaluated in experiments (Sec.~\ref{subsec: abla}).

\vspace{-5pt}
\subsection{Temporal Information Encoding}
\label{subsec: temp_enc}
Dynamic scenes involve complicated point motions in the space. We propose to encode temporal information from two perspectives as follows.

\vspace{-5pt}
\paragraph{Coarse Coordinate Deformation}
Like most previous implicit NeRF methods for dynamic scenes~\cite{pumarola2021d,park2021nerfies,park2021hypernerf}, we introduce a deformation network to shift coordinates of points which simulates the movement of points, but compress the network into a very small one. Denoting coordinates of one point as $(x, y, z)$ and the deformation network as $\Phi_d$, the coordinates are mapped into new ones according to encoded time embeddings $\vt_i = \Phi_t(\gamma(t_i)$:
\begin{equation}
\label{eq: deform}
    x', y', z' = \Phi_d(x, y, z, \vt_i).
\end{equation}

We use only 3-layer MLPs (multilayer perceptrons) as the deformation network, which is much smaller than ones adopted in previous dynamic-scene methods. As this deformation network is applied on every sample point which accounts for a large potion of computation cost, we compress this network from both widths and depths for accelerating optimization and rendering processes.

\vspace{-5pt}
\paragraph{Temporal Information Enhancement}
As the aforementioned deformation network is severely compressed in our method, it may introduces unavoidable deviation for coordinate shifting due to its limited capacity. Besides, this deviation will be aggravated as neural voxels are queried according to point coordinates. Thus, final error not only comes from interpolation weights but also from the queried vertices. As shown in Fig.~\ref{fig: framework}, to alleviate this deviation/mismatch, we propose to further enhance the temporal information by concatenating interpolated features in Eq.~\ref{eq: interp} with positional-encoded coordinates and neural-encoded temporal embeddings. All the concatenated features and embeddings are further fed into neural networks, where the above deviation will be automatically suppressed.

\begin{table*}[thbp]
\newcommand{\xmark}{\ding{55}}
\centering
\caption{Comparisons about training/memory cost and rendering quality on synthetic scenes.}
\Description{Comparisons about training/memory cost and rendering quality on synthetic scenes.}
\vspace{-10pt}
\label{tab: synth-comp}
\begin{tabular}{l|cc|cc|ccc}
\toprule\textbf{Method} & \textbf{w/ Time Enc.} & \textbf{w/ Explicit Rep.} &\textbf{Time}&\textbf{Storage}&\textbf{PSNR} $\uparrow$ & \textbf{SSIM} $\uparrow$ & \textbf{LPIPS} $\downarrow$ \\
\midrule
NeRF~\cite{mildenhall2020nerf} & \xmark & \xmark & $\sim$ hours &  5 MB &19.00&  0.87& 0.18 \\
DirectVoxGO~\cite{sun2021direct} & \xmark & \checkmark & 5 mins & 205 MB & 18.61 & 0.85 & 0.17  \\
Plenoxels~\cite{yu_and_fridovichkeil2021plenoxels} & \xmark & \checkmark & 6 mins & 717 MB & 20.24 & 0.87 & 0.16 \\
\midrule
T-NeRF~\cite{pumarola2021d} & \checkmark & \xmark& $\sim$ hours & -- &29.51 &0.95 & 0.08\\
D-NeRF~\cite{pumarola2021d} & \checkmark & \xmark & 20 hours& 4 MB& 30.50 &0.95 & 0.07\\
\midrule
\name-S (ours) & \checkmark & \checkmark & 8 mins  & 8 MB  &30.75& 0.96&0.07 \\  
 \name-B (ours) & \checkmark &\checkmark & 28 mins & 48 MB & \textbf{32.67} & \textbf{0.97}  & \textbf{0.04} \\
\bottomrule
\end{tabular}
\end{table*}

\begin{figure*}[thbp]
    \vspace{-10pt}
    \centering
    \includegraphics[width=0.95\linewidth]{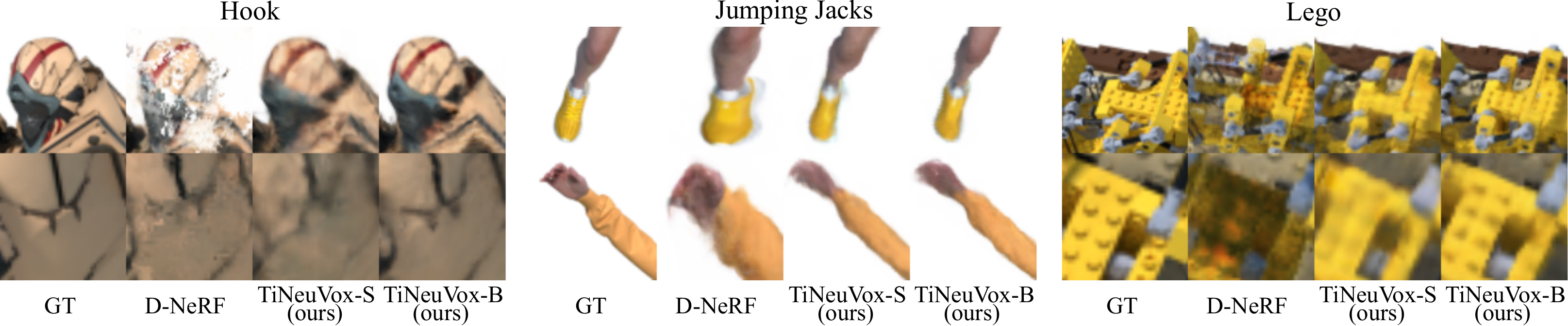}
    \vspace{-12pt}
    \caption{Qualitative comparisons between D-NeRF~\cite{pumarola2021d} and our \name on synthetic scenes.}
    \Description{Qualitative comparisons between D-NeRF and our TiNeuVox on synthetic scenes.}
    \label{fig: syth-imgs}
    \vspace{-10pt}
\end{figure*}

\subsection{Overall Framework and Optimization}
\label{subsec: frame_optim}
Our overall framework is illustrated in Fig.~\ref{fig: framework}. First, the time stamp is encoded by a two-layer MLPs $\Phi_t$, and then fed into a compressed deformation network $\Phi_d$ along with coordinates of the sampled point $(x, y, z)$ to obtain shifted coordinates $(x', y', z')$ as in Eq.~\ref{eq: deform}. The shifted coordinates are used for querying and interpolating neural voxels with a multi-distance manner as in Eq.~\ref{eq: interp}. Then, the concatenated neural voxel \rev{$\vv$} \revb{as in Eq.~\ref{eq: interp}}, encoded time embeddings \revb{$\vt = \Phi_t(\gamma(t)$} and original coordinates $(x, y, z)$ are all concatenated to be fed into a narrow and shallow radiance network $\Phi_r$ to obtain the final density \revb{$\sigma$} and color \revb{$\vc$}:
\begin{equation}
\begin{aligned}
\label{eq: radiance_net}
    \vc,\sigma = \Phi_r(\gamma(\vv), \vt, \gamma(x, y, z), \gamma(\vd)),
\end{aligned}
\end{equation}
where \revb{$\gamma$ denotes the positional encoding as in Eq.~\ref{eq: pe}} and $\vd = (\theta, \phi)$ represents the ray direction which is fed into $\Phi_r$ in the latter stage.

For each ray, we sample points evenly from the near bound to the far bound. By performing the above computation on each sampled point along a ray, the final predicted color can be obtained via volumne rendering as Eq.~\ref{eq: rendering}.
Parameters of all the neural voxels and networks can be optimized by minimizing the distance between predicted colors and groundtruth colors of image pixels as Eq.~\ref{eq: color_loss}. Besides, following \citep{sun2021direct} we adopt two additional loss functions for regularization. One supervises predicted colors of all the sampled points along the ray with the groundtruth image pixel color for stablization. The other one builds a cross-entropy loss on $T_{N+1}$ to distinguish fore- and back-ground, where $T_{N+1}$ denotes the accumulated transmittance for an additional point as computed in Eq.~\ref{eq: rendering}.

Besides, once we obtain the predicted density values with the radiance network, we filter points for the rest part of neural network inference with a pre-defined density threshold. This manner can effectively reduce cost of view-dependent color prediction but rarely affect the rendering quality.

\vspace{-2pt}
\section{Experiments}
In this section, we first provide our implementation details in Sec.~\ref{subsec: implem}. Then we show evaluation results and compare \name with other related methods in Sec.~\ref{subsec: eval}. We further perform a series of ablation studies on the key components of \name and provide detailed results and analysis in Sec.~\ref{subsec: abla}.

\vspace{-7pt}
\subsection{Implementation Details}
\label{subsec: implem}
We implement our framework as Fig.~\ref{fig: framework} mainly with PyTorch~\cite{paszke2019pytorch} and provide two versions, \ie \name-S (small) and \name-B (base). For \name-S, neural voxels are constructed with a resolution of $100^3$ and a channel number of $4$; neural voxels in \name-B are at $160^3\times6$. \rev{All neural voxels are initialized with zero-values.} For acceleration, we set the initial resolutions of voxel grids as $\frac{1}{8}$ of the given ones, which are doubled after $2k$, $4k$, and $6k$ iterations during training. \rev{This manner reduces the training time by 12\% but achieves a similar PSNR.} The channel dimension $C_h$ of hidden layers is set as $64$ for \name-S and $256$ for \name-B. The dimension $C_t$ of time embeddings is set the same as each positional-encoded voxel feature, \ie $20$ for \name-S and $30$ for \name-B. The frequency number $L$ of positional encoding (Eq.~\ref{eq: pe}) is set as $10$ for coordinates $(x,y,z)$, $4$ for view direction $\vd$, $8$ for the time stamp $t$, and $2$ for neural voxels. Points are sampled with a step of half the voxel size along each ray for volume rendering.

\begin{figure*}[thbp]
    \centering
    \includegraphics[width=0.9\linewidth]{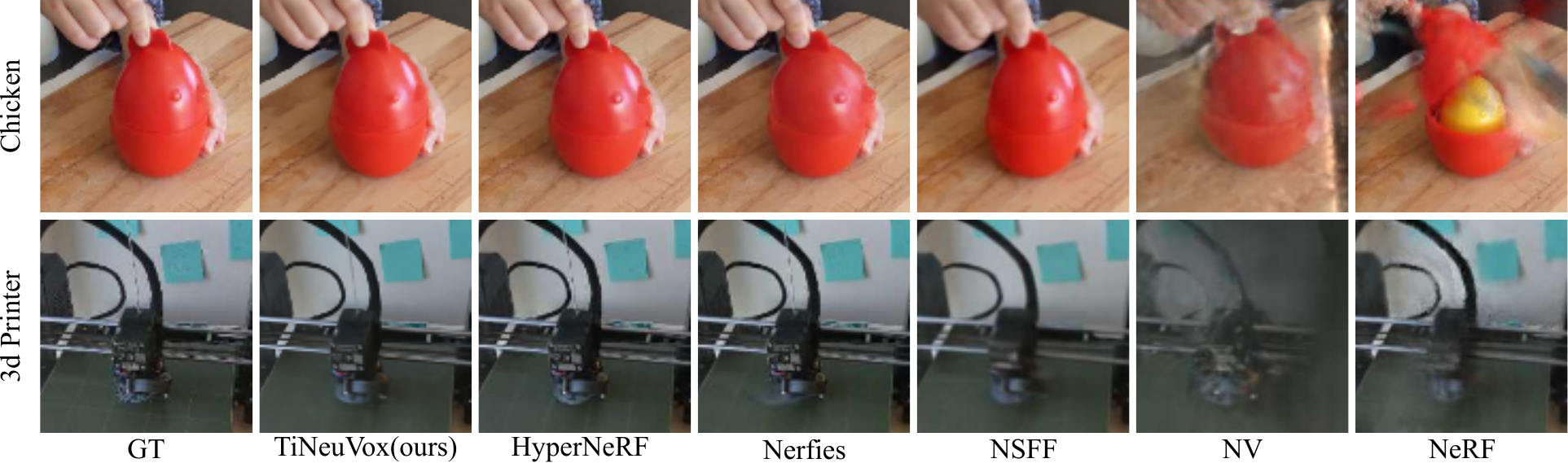}
    \vspace{-10pt}
    \caption{Qualitative comparisons between \name and other methods on real dynamic scenes.}\vspace{-10pt}
    \Description{Qualitative comparisons between TiNeuVox and other methods on real dynamic scenes.}
    \label{fig: real-imgs}
\end{figure*}

For optimization, an Adam~\cite{kingma2015adam} optimizer is used with $(0.9, 0.99)$ $\beta$ values. In each iteration, $4096$ rays are randomly sampled from the whole dataset to form a batch. The initial learning rate is set as $8\times10^{-2}$ for all voxels features, $6\times10^{-4}$ for parameters of the deformation network $\Phi_d$, and $8\times10^{-4}$ for parameters of the other MLPs, which finally decays by $0.1$ with an exponential schedule. The color regularization loss and background cross-entropy loss are weighted by $10^{-2}$ and $10^{-3}$ respectively. To further compress the neural voxel storage, we convert them into the half-precision floating-point format for the last $1k$ iterations. It takes $20k$ iterations in total on one single GeForce RTX 3090 GPU for every scene evaluated in this paper unless specified.

\vspace{-5pt}
\subsection{Evaluation}
\label{subsec: eval}
In this section, we evaluate our method on both synthetic and real dynamic scenes for novel view synthesis. Experimental results are compared with other state-of-the-art (SOTA) methods both quantitatively and qualitatively.

\paragraph{360$^\circ$ Synthetic Scenes}
We adopt the dataset provided by D-NeRF \cite{pumarola2021d} for synthetic-scene evaluation, containing $8$ scenes with dynamic objects under large motions and realistic non-Lambertian materials. Each scene contains $50 - 200$ images for training and $20$ images for testing. To fairly compare with D-NeRF~\cite{pumarola2021d}, each image is trained and rendered at $400 \times 400$ pixels.

\begin{table}[t!]
\setlength{\tabcolsep}{2pt}
\centering
\caption{Quantitative comparisons on real dynamic scenes.}
\Description{Quantitative comparisons on real dynamic scenes.}
\vspace{-8pt}
\label{tab: real-comp}
\begin{tabular}{l|c|cccccccc|cc}
\toprule 
\textbf{Method} & \textbf{Time} & \textbf{PSNR$\uparrow$}  & \textbf{MS-SSIM$\uparrow$} \\
\midrule
    NeRF~\cite{mildenhall2020nerf} & $\sim$ hours & 20.1&0.745 \\
    NV~\cite{lombardi2019neural} & $\sim$ hours & 16.9&0.571 \\
    NSFF~\cite{li2021nsff} & $\sim$ hours & 26.3& 0.916\\
    Nerfies~\cite{park2021nerfies} & $\sim$ hours & 22.2& 0.803\\
    HyperNeRF~\cite{park2021hypernerf} & 32 hours & 22.4 & 0.814\\
    \midrule
    \name-S (ours) & 10 mins & 23.4  & 0.813 \\
    \name-B (ours) &30 mins &24.3  &0.837  \\
\bottomrule
\end{tabular}
\begin{tablenotes}
\footnotesize
\item[] \textdagger \; Time cost of HyperNeRF~\cite{park2021hypernerf} is estimated according to descriptions in their paper but on TPUs.
\vspace{-5pt}
\end{tablenotes}
\end{table}

As shown in Tab.~\ref{tab: synth-comp}, we provide two versions of our \name as described in Sec.~\ref{subsec: implem}. Three metrics are used for evaluation, \ie peak signal-to-noise ratio (PSNR), structural similarity (SSIM)~\cite{wang2004image} and
learned perceptual image patch similarity (LPIPS)~\cite{zhang2018unreasonable}. Conventional NeRF~\cite{mildenhall2020nerf} and the fast-convergence method DirectVoxGO~\cite{sun2021direct} Plenoxels~\cite{yu_and_fridovichkeil2021plenoxels} are all targeted at static scenes, which are unable to model point motions and show bad results. T-NeRF~\cite{pumarola2021d} and D-NeRF~\cite{pumarola2021d} are two dynamic NeRF methods, which model time information via additional input dimension or deformation fields. D-NeRF shows promising rendering quality but it takes about 20 hours on one GPU for per-scene training\footnote{We test this time cost by reproducing D-NeRF~\cite{pumarola2021d} on one single RTX 3090 GPU for fair comparison.}. In comparison, our \name-S takes only $8$ minutes to finish one scene learning and achieves similar rendering performance with three evaluation metrics. For the larger version \name-B, a quite better rendering quality can be achieved ($32.67$ average PSNR) and the optimization can still be finished with only $28$ minutes. It is worth noting that though explicit representations are introduced for acceleration, storage cost of \name is pretty small. \name-S takes only $8$ MB, even similar with pure-implicit D-NeRF; \name-B takes $48$ MB, much smaller than previous explicit methods~\cite{sun2021direct,yu_and_fridovichkeil2021plenoxels}.
Moreover, we provide qualitative comparisons in Fig.~\ref{fig: syth-imgs}, where our \name shows finer and more accurate details than D-NeRF though with much less training time.

\paragraph{Real Scenes}
We further evaluate our method on real non-rigidly deforming scenes provided by HyperNeRF~\cite{park2021hypernerf}. To obtain images from these scenes, a multi-view data capture rig is built with 2 Pixel 3 phones rigidly attached roughly 16cm apart. More details can be referred to in \citet{park2021nerfies,park2021hypernerf}. We perform experiments on four scenes released by \citet{park2021hypernerf}, \ie Broom, 3D Printer, Chicken, and Peel Banana. Following \citet{park2021hypernerf}, PSNR and MS-SSIM~\cite{wang2003multiscale} are used as evaluation metrics\footnote{We find LPIPS~\cite{zhang2018unreasonable} values reported in \cite{park2021hypernerf} hard to reproduce so we omit this evaluation metric.}. Each image is trained and rendered at half of 1080p resolutions, \ie $960 \times 540$ pixels, for quantitative evaluation. To fairly compare with HyperNeRF~\cite{park2021hypernerf}, qualitative results are obtained at full-HD with roughly $1920\times1080$ pixels, taking $40k$ iterations while HyperNeRF takes $1M$ iterations.

As shown in Tab.~\ref{tab: real-comp}, we compare with several highly related neural rendering works towards real dynamic scenes. Taking $192\times$ less training time, our \name achieves similar rendering performance with the previous SOTA method HyperNeRF~\cite{park2021hypernerf}. Noting that though NSFF~\cite{li2021nsff} achieves higher evaluation metrics, our \name and HyperNeRF shows better qualitative rendering quality as in Fig.~\ref{fig: real-imgs}. This phenomenon is also observed in \citet{park2021hypernerf} as quantitative metrics are usually sensitive to small shifts which are yet not obvious to humans. However, our rendered images are slightly more blurred than HyperNeRF. We deduce that far more iterations ($1M$) and additional regularization losses for real dynamic scenes in HyperNeRF matter. We would like to further explore these regularization techniques in future, which are compatible with our frameworks.

\subsection{Ablation Study}
\label{subsec: abla}
In this section, we perform a series of experiments to study key components and factors involved in our method to better understand the mechanism and demonstrate the effectiveness. Following experiments are performed on all the synthetic dynamic scenes \cite{pumarola2021d} and averaged metric values are reported.

\paragraph{Temporal Information Encoding}
We study three components for encoding time information, \ie coordinate deformation $\Phi_d$, temporal information enhancement as in Fig.~\ref{fig: framework}, and encoding time embeddings with the neural network $\Phi_t$. As shown in Tab.~\ref{tab: abla-key-comps}, all of the three components is of critical importance to the final rendering performance.

\begin{table}[t!]
\newcommand{\xmark}{\ding{55}}
\centering
\caption{Ablation study about components of encoding time information with \name-B, \ie deforming coordinates, enhancing temporal information, and encoding neural time embeddings.}
\Description{Ablation study about components for encoding time information.}
\vspace{-10pt}
\setlength{\tabcolsep}{1pt}
\label{tab: abla-key-comps}
\begin{tabular}{ccc|ccc}
\toprule
 \begin{tabular}{c} \textbf{Deform}\\\textbf{Coords.}\end{tabular}& \begin{tabular}{c}\textbf{Enhance}\\\textbf{Temp. Info.} \end{tabular} & 
 \begin{tabular}{c}\textbf{Enc. Neural}\\
\textbf{Time Embeds.}\end{tabular}&
 \textbf{PSNR}  $\uparrow$ & \textbf{SSIM} $\uparrow$ & \textbf{LPIPS} $\downarrow$ \\
\midrule
\checkmark & \checkmark & \checkmark & \textbf{32.668} & \textbf{0.971} & \textbf{0.041} \\
\midrule
\xmark& \checkmark & \checkmark & 29.684& 0.956 & 0.065\\
\midrule
\checkmark &  \xmark &\checkmark & 31.473 &0.968 & 0.045 \\
\midrule
\checkmark & \checkmark &\xmark &  32.384 & 0.971 & 0.044\\
\bottomrule
\end{tabular}
\end{table}

\begin{table}[t!]
\newcommand{\xmark}{\ding{55}}
\centering
\caption{Ablation study about neural voxels with various resolutions and sampling strides for interpolation. ``Res.'' denotes the resolution of neural voxels.}
\Description{Ablation study on neural voxels with various resolutions and sampling strides for interpolation.}
\vspace{-10pt}
\setlength{\tabcolsep}{3pt}
\begin{tabular}{cc|cc|ccc}
\toprule
\textbf{Res.} & \textbf{Strides} & \textbf{Time} & \textbf{Storage} & \textbf{PSNR} $\uparrow$ & \textbf{SSIM} $\uparrow$ & \textbf{LPIPS} $\downarrow$ \\
\midrule
$100^3$ & 1 & 7 mins & 8 MB& 30.279&0.953 &0.072 \\
$100^3$ & 1, 2, 4 &8 mins & 8 MB&30.746& 0.956& 0.067 \\
\midrule
$160^3$ & 1 &10 mins &  32 MB& 30.624& 0.960& 0.072\\
$160^3$ & 1, 2, 4 &12 mins &32 MB & 31.498 & 0.964 & 0.062 \\
\midrule
$256^3$ & 1 &15 mins&  128 MB&30.100& 0.961& 0.082 \\
$256^3$ & 1, 2, 4 & 21 mins& 128 MB&31.693 & 0.968 & 0.061  \\
$256^3$ & 1, 2, 4, 8 & 23 mins& 128 MB & 31.792 & 0.969 & 0.056 \\
\bottomrule
\end{tabular}
\vspace{-3pt}
\label{tab: abla-interp}
\end{table}

\paragraph{Multi-distance Interpolation Effectiveness}
We study effectiveness of the proposed multi-distance interpolation (MDI) and show results in Tab.~\ref{tab: abla-interp}. Experiments with a single $1$ sampling stride equal to interpolation without the multi-distance manner. For the small resolution setting with $100^3$, MDI brings a $0.467$ PSNR promotion. It can be observed the larger resolutions are, the bigger advantages MDI bring, \ie $0.874$ for $160^3$ and $1.593$ for $256^3$. This is because larger the resolution is, a smaller region each grid can represent. Noting that for $256^3$, each grid is too small to capture complete motions without MDI. 

\begin{figure}[t!]
    \centering
    \includegraphics[width=\linewidth]{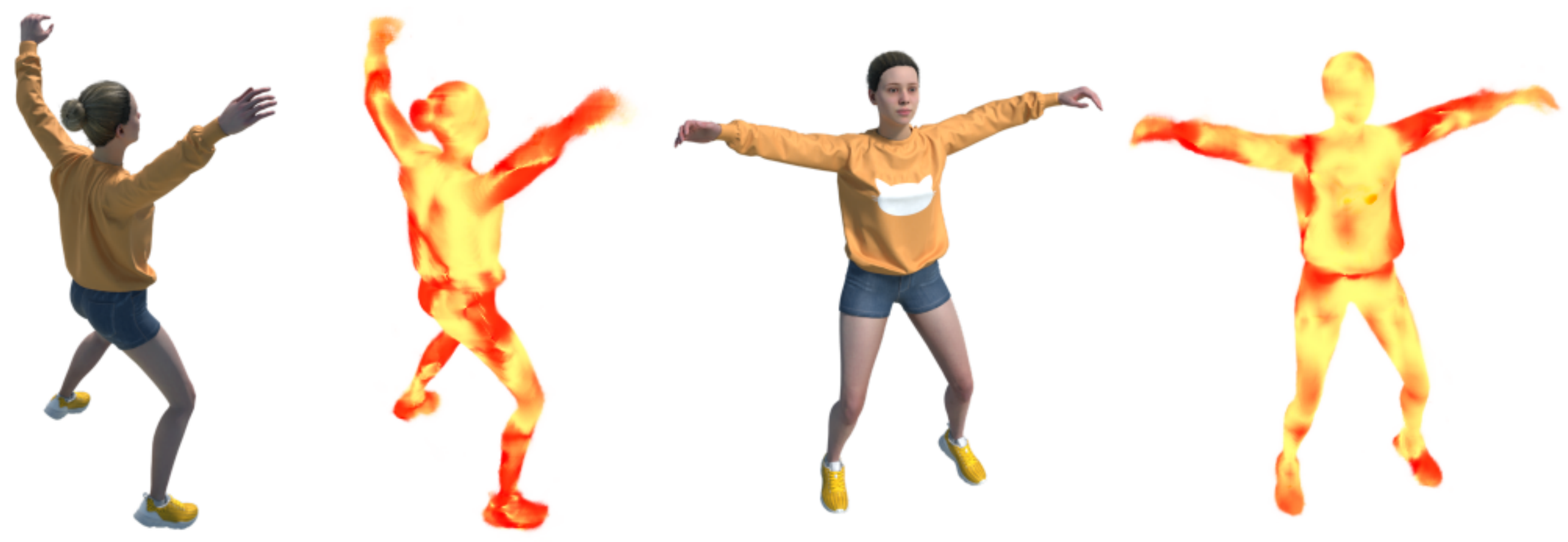}
    \vspace{-15pt}
    \caption{\rev{Visualization of gradient magnitudes on neural voxels with different interpolation distances. Red colors denote voxels from longer distances have larger gradient magnitudes, while yellow ones indicate nearer voxels.}}
    \Description{Visualization of gradient magnitudes on neural voxels with different interpolation distances.}
    \label{fig: vox-grad}
\end{figure}

To clearly demonstrate the MDI mechanism, we visualize gradient magnitudes on neural voxels interpolated with different distances, where red colors denote distant voxel samples have larger gradients and yellow colors denote near samples have larger gradients. Similar to ideas in Grad-CAM~\cite{selvaraju2017grad}, gradient magnitudes represent the activation strength or importance of voxel samples. As shown in Fig.~\ref{fig: vox-grad}, distant voxels samples (\textcolor{red}{red}) show higher importance for large-motion points (\eg arms); small-motion points (\eg head) prefer near voxel samples (\textcolor{amber}{yellow}).
We clarify the MDI mechanism as follows. The coordinate deformation network is highly compressed so it can only model a coarse motion. This deviation becomes more dominant for larger motions and may lead to wrong voxel queries. This exacerbates even more at higher-resolution voxel grids. MDI enables each voxel to perceive multi-distance points; hence even when wrong voxels are queried, voxel features can still provide correct information.

\section{Discussion and Conclusion}
\rev{
\paragraph{Limitations \& Future Works}
We perform preliminary experiments on the NSFF~\cite{li2021nsff} scene "truck", which contains a long-distance motion. Our method produces plausible quality but there are still several opportunities for future research, including consideration of relations between neighboring frames and leveraging spatial/temporal partitioning of neural networks; \eg, voxels can be divided into blocks that are tracked over time. Handling specularities is also a generally difficult setting, where our approach has limitations in common with existing neural representation works. One potential future avenue is to specifically address reflections as shown in Ref-NeRF~\cite{verbin2022refnerf}, which however is orthogonal to goals of our work. \revb{There still remain} some other critical topics, \eg regularization techniques for complicated real scenes, integrating geometric/motion priors for domain-specific scenes, and further introducing compressing/pruning techniques like voxel hashing~\cite{niessner2013real,muller2022instant} and tensor decomposition~\cite{chen2022tensorf} \etc.}

In this paper, we propose a fast neural rendering framework \name targeted at dynamic scenes. Different from static scenes, dynamic ones involve complex point/object motions. We construct time-aware neural voxels to represent scenes, where temporal information is encoded with a highly compressed deformation network and enhanced in the radiance network. A multi-distance interpolation method is proposed to model accurate motions of various scales. \name can achieve extremely fast training speed with low storage cost. How to equip pure-spatial neural voxels with temporal information is an interesting and valuable question. We believe our work can shed light on neural rendering acceleration for dynamic scenes.

\begin{acks}
We thank Prof. Angela Dai for narration in the video, and Liangchen Song and Yingqing Rao for their valuable discussions and assistance. This work was supported by National Natural Science Foundation of China (NSFC No. 61733007 and No. 61876212).
\end{acks}

\bibliographystyle{ACM-Reference-Format}
\bibliography{sample-base}


\begin{thebibliography}{56}


\ifx \showCODEN    \undefined \def \showCODEN     #1{\unskip}     \fi
\ifx \showDOI      \undefined \def \showDOI       #1{#1}\fi
\ifx \showISBNx    \undefined \def \showISBNx     #1{\unskip}     \fi
\ifx \showISBNxiii \undefined \def \showISBNxiii  #1{\unskip}     \fi
\ifx \showISSN     \undefined \def \showISSN      #1{\unskip}     \fi
\ifx \showLCCN     \undefined \def \showLCCN      #1{\unskip}     \fi
\ifx \shownote     \undefined \def \shownote      #1{#1}          \fi
\ifx \showarticletitle \undefined \def \showarticletitle #1{#1}   \fi
\ifx \showURL      \undefined \def \showURL       {\relax}        \fi
\providecommand\bibfield[2]{#2}
\providecommand\bibinfo[2]{#2}
\providecommand\natexlab[1]{#1}
\providecommand\showeprint[2][]{arXiv:#2}

\bibitem[\protect\citeauthoryear{Arandjelovi{\'c} and
  Zisserman}{Arandjelovi{\'c} and Zisserman}{2021}]%
        {arandjelovic2021nerf}
\bibfield{author}{\bibinfo{person}{Relja Arandjelovi{\'c}} {and}
  \bibinfo{person}{Andrew Zisserman}.} \bibinfo{year}{2021}\natexlab{}.
\newblock \showarticletitle{NeRF in detail: Learning to sample for view
  synthesis}.
\newblock \bibinfo{journal}{\emph{arXiv:2106.05264}} (\bibinfo{year}{2021}).
\newblock


\bibitem[\protect\citeauthoryear{Attal, Laidlaw, Gokaslan, Kim, Richardt,
  Tompkin, and O'Toole}{Attal et~al\mbox{.}}{2021}]%
        {attal2021torf}
\bibfield{author}{\bibinfo{person}{Benjamin Attal}, \bibinfo{person}{Eliot
  Laidlaw}, \bibinfo{person}{Aaron Gokaslan}, \bibinfo{person}{Changil Kim},
  \bibinfo{person}{Christian Richardt}, \bibinfo{person}{James Tompkin}, {and}
  \bibinfo{person}{Matthew O'Toole}.} \bibinfo{year}{2021}\natexlab{}.
\newblock \showarticletitle{T{\"o}rf: Time-of-flight radiance fields for
  dynamic scene view synthesis}.
\newblock \bibinfo{journal}{\emph{NeurIPS}} (\bibinfo{year}{2021}).
\newblock


\bibitem[\protect\citeauthoryear{Barron, Mildenhall, Tancik, Hedman,
  Martin-Brualla, and Srinivasan}{Barron et~al\mbox{.}}{2021}]%
        {Barron_2021_ICCV}
\bibfield{author}{\bibinfo{person}{Jonathan~T. Barron}, \bibinfo{person}{Ben
  Mildenhall}, \bibinfo{person}{Matthew Tancik}, \bibinfo{person}{Peter
  Hedman}, \bibinfo{person}{Ricardo Martin-Brualla}, {and}
  \bibinfo{person}{Pratul~P. Srinivasan}.} \bibinfo{year}{2021}\natexlab{}.
\newblock \showarticletitle{Mip-NeRF: A Multiscale Representation for
  Anti-Aliasing Neural Radiance Fields}. In \bibinfo{booktitle}{\emph{ICCV}}.
\newblock


\bibitem[\protect\citeauthoryear{Chen, Xu, Geiger, Yu, and Su}{Chen
  et~al\mbox{.}}{2022}]%
        {chen2022tensorf}
\bibfield{author}{\bibinfo{person}{Anpei Chen}, \bibinfo{person}{Zexiang Xu},
  \bibinfo{person}{Andreas Geiger}, \bibinfo{person}{Jingyi Yu}, {and}
  \bibinfo{person}{Hao Su}.} \bibinfo{year}{2022}\natexlab{}.
\newblock \showarticletitle{TensoRF: Tensorial Radiance Fields}.
\newblock \bibinfo{journal}{\emph{arXiv:2203.09517}} (\bibinfo{year}{2022}).
\newblock


\bibitem[\protect\citeauthoryear{Chen, Xu, Zhao, Zhang, Xiang, Yu, and Su}{Chen
  et~al\mbox{.}}{2021}]%
        {Chen_2021_ICCV}
\bibfield{author}{\bibinfo{person}{Anpei Chen}, \bibinfo{person}{Zexiang Xu},
  \bibinfo{person}{Fuqiang Zhao}, \bibinfo{person}{Xiaoshuai Zhang},
  \bibinfo{person}{Fanbo Xiang}, \bibinfo{person}{Jingyi Yu}, {and}
  \bibinfo{person}{Hao Su}.} \bibinfo{year}{2021}\natexlab{}.
\newblock \showarticletitle{MVSNeRF: Fast Generalizable Radiance Field
  Reconstruction From Multi-View Stereo}. In \bibinfo{booktitle}{\emph{ICCV}}.
\newblock


\bibitem[\protect\citeauthoryear{Deng, Liu, Zhu, and Ramanan}{Deng
  et~al\mbox{.}}{2021}]%
        {deng2021depth}
\bibfield{author}{\bibinfo{person}{Kangle Deng}, \bibinfo{person}{Andrew Liu},
  \bibinfo{person}{Jun-Yan Zhu}, {and} \bibinfo{person}{Deva Ramanan}.}
  \bibinfo{year}{2021}\natexlab{}.
\newblock \showarticletitle{Depth-supervised nerf: Fewer views and faster
  training for free}.
\newblock \bibinfo{journal}{\emph{arXiv:2107.02791}} (\bibinfo{year}{2021}).
\newblock


\bibitem[\protect\citeauthoryear{Du, Zhang, Yu, Tenenbaum, and Wu}{Du
  et~al\mbox{.}}{2021}]%
        {du2021neural}
\bibfield{author}{\bibinfo{person}{Yilun Du}, \bibinfo{person}{Yinan Zhang},
  \bibinfo{person}{Hong-Xing Yu}, \bibinfo{person}{Joshua~B Tenenbaum}, {and}
  \bibinfo{person}{Jiajun Wu}.} \bibinfo{year}{2021}\natexlab{}.
\newblock \showarticletitle{Neural radiance flow for 4d view synthesis and
  video processing}. In \bibinfo{booktitle}{\emph{ICCV}}.
\newblock


\bibitem[\protect\citeauthoryear{Fang, Xie, Wang, Zhang, Liu, and Tian}{Fang
  et~al\mbox{.}}{2021}]%
        {fang2021neusample}
\bibfield{author}{\bibinfo{person}{Jiemin Fang}, \bibinfo{person}{Lingxi Xie},
  \bibinfo{person}{Xinggang Wang}, \bibinfo{person}{Xiaopeng Zhang},
  \bibinfo{person}{Wenyu Liu}, {and} \bibinfo{person}{Qi Tian}.}
  \bibinfo{year}{2021}\natexlab{}.
\newblock \showarticletitle{NeuSample: Neural Sample Field for Efficient View
  Synthesis}.
\newblock \bibinfo{journal}{\emph{arXiv:2111.15552}} (\bibinfo{year}{2021}).
\newblock


\bibitem[\protect\citeauthoryear{Gao, Saraf, Kopf, and Huang}{Gao
  et~al\mbox{.}}{2021}]%
        {gao2021dynamic}
\bibfield{author}{\bibinfo{person}{Chen Gao}, \bibinfo{person}{Ayush Saraf},
  \bibinfo{person}{Johannes Kopf}, {and} \bibinfo{person}{Jia-Bin Huang}.}
  \bibinfo{year}{2021}\natexlab{}.
\newblock \showarticletitle{Dynamic view synthesis from dynamic monocular
  video}. In \bibinfo{booktitle}{\emph{ICCV}}.
\newblock


\bibitem[\protect\citeauthoryear{Garbin, Kowalski, Johnson, Shotton, and
  Valentin}{Garbin et~al\mbox{.}}{2021}]%
        {Garbin_2021_ICCV}
\bibfield{author}{\bibinfo{person}{Stephan~J. Garbin}, \bibinfo{person}{Marek
  Kowalski}, \bibinfo{person}{Matthew Johnson}, \bibinfo{person}{Jamie
  Shotton}, {and} \bibinfo{person}{Julien Valentin}.}
  \bibinfo{year}{2021}\natexlab{}.
\newblock \showarticletitle{FastNeRF: High-Fidelity Neural Rendering at
  200FPS}. In \bibinfo{booktitle}{\emph{ICCV}}.
\newblock


\bibitem[\protect\citeauthoryear{Hedman, Srinivasan, Mildenhall, Barron, and
  Debevec}{Hedman et~al\mbox{.}}{2021}]%
        {Hedman_2021_ICCV}
\bibfield{author}{\bibinfo{person}{Peter Hedman}, \bibinfo{person}{Pratul~P.
  Srinivasan}, \bibinfo{person}{Ben Mildenhall}, \bibinfo{person}{Jonathan~T.
  Barron}, {and} \bibinfo{person}{Paul Debevec}.}
  \bibinfo{year}{2021}\natexlab{}.
\newblock \showarticletitle{Baking Neural Radiance Fields for Real-Time View
  Synthesis}. In \bibinfo{booktitle}{\emph{ICCV}}.
\newblock


\bibitem[\protect\citeauthoryear{Kajiya and Von~Herzen}{Kajiya and
  Von~Herzen}{1984}]%
        {kajiya1984ray}
\bibfield{author}{\bibinfo{person}{James~T Kajiya} {and}
  \bibinfo{person}{Brian~P Von~Herzen}.} \bibinfo{year}{1984}\natexlab{}.
\newblock \showarticletitle{Ray tracing volume densities}.
\newblock \bibinfo{journal}{\emph{ACM SIGGRAPH computer graphics}}
  (\bibinfo{year}{1984}).
\newblock


\bibitem[\protect\citeauthoryear{Kingma and Ba}{Kingma and Ba}{2015}]%
        {kingma2015adam}
\bibfield{author}{\bibinfo{person}{Diederik~P Kingma} {and}
  \bibinfo{person}{Jimmy Ba}.} \bibinfo{year}{2015}\natexlab{}.
\newblock \showarticletitle{Adam: A Method for Stochastic Optimization}. In
  \bibinfo{booktitle}{\emph{ICLR}}.
\newblock


\bibitem[\protect\citeauthoryear{Kolda and Bader}{Kolda and Bader}{2009}]%
        {kolda2009tensor}
\bibfield{author}{\bibinfo{person}{Tamara~G Kolda} {and}
  \bibinfo{person}{Brett~W Bader}.} \bibinfo{year}{2009}\natexlab{}.
\newblock \showarticletitle{Tensor decompositions and applications}.
\newblock \bibinfo{journal}{\emph{SIAM review}} (\bibinfo{year}{2009}).
\newblock


\bibitem[\protect\citeauthoryear{Li, Slavcheva, Zollhoefer, Green, Lassner,
  Kim, Schmidt, Lovegrove, Goesele, and Lv}{Li et~al\mbox{.}}{2021b}]%
        {li2021neural}
\bibfield{author}{\bibinfo{person}{Tianye Li}, \bibinfo{person}{Mira
  Slavcheva}, \bibinfo{person}{Michael Zollhoefer}, \bibinfo{person}{Simon
  Green}, \bibinfo{person}{Christoph Lassner}, \bibinfo{person}{Changil Kim},
  \bibinfo{person}{Tanner Schmidt}, \bibinfo{person}{Steven Lovegrove},
  \bibinfo{person}{Michael Goesele}, {and} \bibinfo{person}{Zhaoyang Lv}.}
  \bibinfo{year}{2021}\natexlab{b}.
\newblock \showarticletitle{Neural 3d video synthesis}.
\newblock \bibinfo{journal}{\emph{arXiv:2103.02597}} (\bibinfo{year}{2021}).
\newblock


\bibitem[\protect\citeauthoryear{Li, Niklaus, Snavely, and Wang}{Li
  et~al\mbox{.}}{2021a}]%
        {li2021nsff}
\bibfield{author}{\bibinfo{person}{Zhengqi Li}, \bibinfo{person}{Simon
  Niklaus}, \bibinfo{person}{Noah Snavely}, {and} \bibinfo{person}{Oliver
  Wang}.} \bibinfo{year}{2021}\natexlab{a}.
\newblock \showarticletitle{Neural scene flow fields for space-time view
  synthesis of dynamic scenes}. In \bibinfo{booktitle}{\emph{CVPR}}.
\newblock


\bibitem[\protect\citeauthoryear{Lin, Ma, Torralba, and Lucey}{Lin
  et~al\mbox{.}}{2021}]%
        {lin2021barf}
\bibfield{author}{\bibinfo{person}{Chen-Hsuan Lin}, \bibinfo{person}{Wei-Chiu
  Ma}, \bibinfo{person}{Antonio Torralba}, {and} \bibinfo{person}{Simon
  Lucey}.} \bibinfo{year}{2021}\natexlab{}.
\newblock \showarticletitle{Barf: Bundle-adjusting neural radiance fields}. In
  \bibinfo{booktitle}{\emph{ICCV}}.
\newblock


\bibitem[\protect\citeauthoryear{Lindell, Martel, and Wetzstein}{Lindell
  et~al\mbox{.}}{2021}]%
        {lindell2021autoint}
\bibfield{author}{\bibinfo{person}{David~B Lindell}, \bibinfo{person}{Julien~NP
  Martel}, {and} \bibinfo{person}{Gordon Wetzstein}.}
  \bibinfo{year}{2021}\natexlab{}.
\newblock \showarticletitle{Autoint: Automatic integration for fast neural
  volume rendering}. In \bibinfo{booktitle}{\emph{CVPR}}.
\newblock


\bibitem[\protect\citeauthoryear{Liu, Gu, Lin, Chua, and Theobalt}{Liu
  et~al\mbox{.}}{2020}]%
        {LiuGLCT20}
\bibfield{author}{\bibinfo{person}{Lingjie Liu}, \bibinfo{person}{Jiatao Gu},
  \bibinfo{person}{Kyaw~Zaw Lin}, \bibinfo{person}{Tat{-}Seng Chua}, {and}
  \bibinfo{person}{Christian Theobalt}.} \bibinfo{year}{2020}\natexlab{}.
\newblock \showarticletitle{Neural Sparse Voxel Fields}. In
  \bibinfo{booktitle}{\emph{NeurIPS}}.
\newblock


\bibitem[\protect\citeauthoryear{Liu, Peng, Liu, Wang, Wang, Christian, Zhou,
  and Wang}{Liu et~al\mbox{.}}{2022}]%
        {liu2022neuray}
\bibfield{author}{\bibinfo{person}{Yuan Liu}, \bibinfo{person}{Sida Peng},
  \bibinfo{person}{Lingjie Liu}, \bibinfo{person}{Qianqian Wang},
  \bibinfo{person}{Peng Wang}, \bibinfo{person}{Theobalt Christian},
  \bibinfo{person}{Xiaowei Zhou}, {and} \bibinfo{person}{Wenping Wang}.}
  \bibinfo{year}{2022}\natexlab{}.
\newblock \showarticletitle{Neural Rays for Occlusion-aware Image-based
  Rendering}. In \bibinfo{booktitle}{\emph{CVPR}}.
\newblock


\bibitem[\protect\citeauthoryear{Lombardi, Simon, Saragih, Schwartz, Lehrmann,
  and Sheikh}{Lombardi et~al\mbox{.}}{2019}]%
        {lombardi2019neural}
\bibfield{author}{\bibinfo{person}{Stephen Lombardi}, \bibinfo{person}{Tomas
  Simon}, \bibinfo{person}{Jason Saragih}, \bibinfo{person}{Gabriel Schwartz},
  \bibinfo{person}{Andreas Lehrmann}, {and} \bibinfo{person}{Yaser Sheikh}.}
  \bibinfo{year}{2019}\natexlab{}.
\newblock \showarticletitle{Neural volumes: learning dynamic renderable volumes
  from images}.
\newblock \bibinfo{journal}{\emph{ACM Transactions on Graphics}}
  (\bibinfo{year}{2019}).
\newblock


\bibitem[\protect\citeauthoryear{Martin-Brualla, Radwan, Sajjadi, Barron,
  Dosovitskiy, and Duckworth}{Martin-Brualla et~al\mbox{.}}{2021}]%
        {Martin-Brualla_2021_CVPR}
\bibfield{author}{\bibinfo{person}{Ricardo Martin-Brualla},
  \bibinfo{person}{Noha Radwan}, \bibinfo{person}{Mehdi S.~M. Sajjadi},
  \bibinfo{person}{Jonathan~T. Barron}, \bibinfo{person}{Alexey Dosovitskiy},
  {and} \bibinfo{person}{Daniel Duckworth}.} \bibinfo{year}{2021}\natexlab{}.
\newblock \showarticletitle{NeRF in the Wild: Neural Radiance Fields for
  Unconstrained Photo Collections}. In \bibinfo{booktitle}{\emph{CVPR}}.
\newblock


\bibitem[\protect\citeauthoryear{Mildenhall, Srinivasan, Tancik, Barron,
  Ramamoorthi, and Ng}{Mildenhall et~al\mbox{.}}{2020}]%
        {mildenhall2020nerf}
\bibfield{author}{\bibinfo{person}{Ben Mildenhall}, \bibinfo{person}{Pratul~P
  Srinivasan}, \bibinfo{person}{Matthew Tancik}, \bibinfo{person}{Jonathan~T
  Barron}, \bibinfo{person}{Ravi Ramamoorthi}, {and} \bibinfo{person}{Ren Ng}.}
  \bibinfo{year}{2020}\natexlab{}.
\newblock \showarticletitle{Nerf: Representing scenes as neural radiance fields
  for view synthesis}. In \bibinfo{booktitle}{\emph{ECCV}}.
\newblock


\bibitem[\protect\citeauthoryear{M{\"u}ller, Evans, Schied, and
  Keller}{M{\"u}ller et~al\mbox{.}}{2022}]%
        {muller2022instant}
\bibfield{author}{\bibinfo{person}{Thomas M{\"u}ller}, \bibinfo{person}{Alex
  Evans}, \bibinfo{person}{Christoph Schied}, {and} \bibinfo{person}{Alexander
  Keller}.} \bibinfo{year}{2022}\natexlab{}.
\newblock \showarticletitle{Instant Neural Graphics Primitives with a
  Multiresolution Hash Encoding}.
\newblock \bibinfo{journal}{\emph{arXiv:2201.05989}} (\bibinfo{year}{2022}).
\newblock


\bibitem[\protect\citeauthoryear{Neff, Stadlbauer, Parger, Kurz, Mueller,
  Chaitanya, Kaplanyan, and Steinberger}{Neff et~al\mbox{.}}{2021}]%
        {neff2021donerf}
\bibfield{author}{\bibinfo{person}{Thomas Neff}, \bibinfo{person}{Pascal
  Stadlbauer}, \bibinfo{person}{Mathias Parger}, \bibinfo{person}{Andreas
  Kurz}, \bibinfo{person}{Joerg~H. Mueller}, \bibinfo{person}{Chakravarty
  R.~Alla Chaitanya}, \bibinfo{person}{Anton~S. Kaplanyan}, {and}
  \bibinfo{person}{Markus Steinberger}.} \bibinfo{year}{2021}\natexlab{}.
\newblock \showarticletitle{{DONeRF: Towards Real-Time Rendering of Compact
  Neural Radiance Fields using Depth Oracle Networks}}.
\newblock \bibinfo{journal}{\emph{Computer Graphics Forum}}
  (\bibinfo{year}{2021}).
\newblock


\bibitem[\protect\citeauthoryear{Nie{\ss}ner, Zollh{\"o}fer, Izadi, and
  Stamminger}{Nie{\ss}ner et~al\mbox{.}}{2013}]%
        {niessner2013real}
\bibfield{author}{\bibinfo{person}{Matthias Nie{\ss}ner},
  \bibinfo{person}{Michael Zollh{\"o}fer}, \bibinfo{person}{Shahram Izadi},
  {and} \bibinfo{person}{Marc Stamminger}.} \bibinfo{year}{2013}\natexlab{}.
\newblock \showarticletitle{Real-time 3D reconstruction at scale using voxel
  hashing}.
\newblock \bibinfo{journal}{\emph{ACM Transactions on Graphics (ToG)}}
  (\bibinfo{year}{2013}).
\newblock


\bibitem[\protect\citeauthoryear{Noguchi, Sun, Lin, and Harada}{Noguchi
  et~al\mbox{.}}{2021}]%
        {noguchi2021neural}
\bibfield{author}{\bibinfo{person}{Atsuhiro Noguchi}, \bibinfo{person}{Xiao
  Sun}, \bibinfo{person}{Stephen Lin}, {and} \bibinfo{person}{Tatsuya Harada}.}
  \bibinfo{year}{2021}\natexlab{}.
\newblock \showarticletitle{Neural articulated radiance field}. In
  \bibinfo{booktitle}{\emph{ICCV}}.
\newblock


\bibitem[\protect\citeauthoryear{Park, Sinha, Barron, Bouaziz, Goldman, Seitz,
  and Martin-Brualla}{Park et~al\mbox{.}}{2021a}]%
        {park2021nerfies}
\bibfield{author}{\bibinfo{person}{Keunhong Park}, \bibinfo{person}{Utkarsh
  Sinha}, \bibinfo{person}{Jonathan~T. Barron}, \bibinfo{person}{Sofien
  Bouaziz}, \bibinfo{person}{Dan~B Goldman}, \bibinfo{person}{Steven~M. Seitz},
  {and} \bibinfo{person}{Ricardo Martin-Brualla}.}
  \bibinfo{year}{2021}\natexlab{a}.
\newblock \showarticletitle{Nerfies: Deformable Neural Radiance Fields}.
\newblock \bibinfo{journal}{\emph{ICCV}} (\bibinfo{year}{2021}).
\newblock


\bibitem[\protect\citeauthoryear{Park, Sinha, Hedman, Barron, Bouaziz, Goldman,
  Martin-Brualla, and Seitz}{Park et~al\mbox{.}}{2021b}]%
        {park2021hypernerf}
\bibfield{author}{\bibinfo{person}{Keunhong Park}, \bibinfo{person}{Utkarsh
  Sinha}, \bibinfo{person}{Peter Hedman}, \bibinfo{person}{Jonathan~T. Barron},
  \bibinfo{person}{Sofien Bouaziz}, \bibinfo{person}{Dan~B Goldman},
  \bibinfo{person}{Ricardo Martin-Brualla}, {and} \bibinfo{person}{Steven~M.
  Seitz}.} \bibinfo{year}{2021}\natexlab{b}.
\newblock \showarticletitle{HyperNeRF: A Higher-Dimensional Representation for
  Topologically Varying Neural Radiance Fields}.
\newblock \bibinfo{journal}{\emph{ACM Trans. Graph.}} (\bibinfo{year}{2021}).
\newblock


\bibitem[\protect\citeauthoryear{Paszke, Gross, Massa, Lerer, Bradbury, Chanan,
  Killeen, Lin, Gimelshein, Antiga, et~al\mbox{.}}{Paszke
  et~al\mbox{.}}{2019}]%
        {paszke2019pytorch}
\bibfield{author}{\bibinfo{person}{Adam Paszke}, \bibinfo{person}{Sam Gross},
  \bibinfo{person}{Francisco Massa}, \bibinfo{person}{Adam Lerer},
  \bibinfo{person}{James Bradbury}, \bibinfo{person}{Gregory Chanan},
  \bibinfo{person}{Trevor Killeen}, \bibinfo{person}{Zeming Lin},
  \bibinfo{person}{Natalia Gimelshein}, \bibinfo{person}{Luca Antiga},
  {et~al\mbox{.}}} \bibinfo{year}{2019}\natexlab{}.
\newblock \showarticletitle{Pytorch: An imperative style, high-performance deep
  learning library}.
\newblock \bibinfo{journal}{\emph{NeurIPS}} (\bibinfo{year}{2019}).
\newblock


\bibitem[\protect\citeauthoryear{Piala and Clark}{Piala and Clark}{2021}]%
        {piala2021terminerf}
\bibfield{author}{\bibinfo{person}{Martin Piala} {and} \bibinfo{person}{Ronald
  Clark}.} \bibinfo{year}{2021}\natexlab{}.
\newblock \showarticletitle{TermiNeRF: Ray Termination Prediction for Efficient
  Neural Rendering}. In \bibinfo{booktitle}{\emph{3DV}}.
\newblock


\bibitem[\protect\citeauthoryear{Pumarola, Corona, Pons-Moll, and
  Moreno-Noguer}{Pumarola et~al\mbox{.}}{2021}]%
        {pumarola2021d}
\bibfield{author}{\bibinfo{person}{Albert Pumarola}, \bibinfo{person}{Enric
  Corona}, \bibinfo{person}{Gerard Pons-Moll}, {and} \bibinfo{person}{Francesc
  Moreno-Noguer}.} \bibinfo{year}{2021}\natexlab{}.
\newblock \showarticletitle{D-nerf: Neural radiance fields for dynamic scenes}.
  In \bibinfo{booktitle}{\emph{CVPR}}.
\newblock


\bibitem[\protect\citeauthoryear{Reiser, Peng, Liao, and Geiger}{Reiser
  et~al\mbox{.}}{2021}]%
        {Reiser_2021_ICCV}
\bibfield{author}{\bibinfo{person}{Christian Reiser}, \bibinfo{person}{Songyou
  Peng}, \bibinfo{person}{Yiyi Liao}, {and} \bibinfo{person}{Andreas Geiger}.}
  \bibinfo{year}{2021}\natexlab{}.
\newblock \showarticletitle{KiloNeRF: Speeding Up Neural Radiance Fields With
  Thousands of Tiny MLPs}. In \bibinfo{booktitle}{\emph{ICCV}}.
\newblock


\bibitem[\protect\citeauthoryear{Selvaraju, Cogswell, Das, Vedantam, Parikh,
  and Batra}{Selvaraju et~al\mbox{.}}{2017}]%
        {selvaraju2017grad}
\bibfield{author}{\bibinfo{person}{Ramprasaath~R Selvaraju},
  \bibinfo{person}{Michael Cogswell}, \bibinfo{person}{Abhishek Das},
  \bibinfo{person}{Ramakrishna Vedantam}, \bibinfo{person}{Devi Parikh}, {and}
  \bibinfo{person}{Dhruv Batra}.} \bibinfo{year}{2017}\natexlab{}.
\newblock \showarticletitle{Grad-cam: Visual explanations from deep networks
  via gradient-based localization}. In \bibinfo{booktitle}{\emph{ICCV}}.
\newblock


\bibitem[\protect\citeauthoryear{Sitzmann, Rezchikov, Freeman, Tenenbaum, and
  Durand}{Sitzmann et~al\mbox{.}}{2021}]%
        {sitzmann2021lfns}
\bibfield{author}{\bibinfo{person}{Vincent Sitzmann}, \bibinfo{person}{Semon
  Rezchikov}, \bibinfo{person}{William~T. Freeman}, \bibinfo{person}{Joshua~B.
  Tenenbaum}, {and} \bibinfo{person}{Fredo Durand}.}
  \bibinfo{year}{2021}\natexlab{}.
\newblock \showarticletitle{Light Field Networks: Neural Scene Representations
  with Single-Evaluation Rendering}. In \bibinfo{booktitle}{\emph{NeurIPS}}.
\newblock


\bibitem[\protect\citeauthoryear{Sitzmann, Thies, Heide, Nie{\ss}ner,
  Wetzstein, and Zollhofer}{Sitzmann et~al\mbox{.}}{2019}]%
        {sitzmann2019deepvoxels}
\bibfield{author}{\bibinfo{person}{Vincent Sitzmann}, \bibinfo{person}{Justus
  Thies}, \bibinfo{person}{Felix Heide}, \bibinfo{person}{Matthias
  Nie{\ss}ner}, \bibinfo{person}{Gordon Wetzstein}, {and}
  \bibinfo{person}{Michael Zollhofer}.} \bibinfo{year}{2019}\natexlab{}.
\newblock \showarticletitle{Deepvoxels: Learning persistent 3d feature
  embeddings}. In \bibinfo{booktitle}{\emph{CVPR}}.
\newblock


\bibitem[\protect\citeauthoryear{Su, Yu, Zollh{\"o}fer, and Rhodin}{Su
  et~al\mbox{.}}{2021}]%
        {su2021nerf}
\bibfield{author}{\bibinfo{person}{Shih-Yang Su}, \bibinfo{person}{Frank Yu},
  \bibinfo{person}{Michael Zollh{\"o}fer}, {and} \bibinfo{person}{Helge
  Rhodin}.} \bibinfo{year}{2021}\natexlab{}.
\newblock \showarticletitle{A-nerf: Articulated neural radiance fields for
  learning human shape, appearance, and pose}.
\newblock \bibinfo{journal}{\emph{NeurIPS}} (\bibinfo{year}{2021}).
\newblock


\bibitem[\protect\citeauthoryear{Sun, Sun, and Chen}{Sun et~al\mbox{.}}{2022}]%
        {sun2021direct}
\bibfield{author}{\bibinfo{person}{Cheng Sun}, \bibinfo{person}{Min Sun}, {and}
  \bibinfo{person}{Hwann-Tzong Chen}.} \bibinfo{year}{2022}\natexlab{}.
\newblock \showarticletitle{Direct Voxel Grid Optimization: Super-fast
  Convergence for Radiance Fields Reconstruction}. In
  \bibinfo{booktitle}{\emph{CVPR}}.
\newblock


\bibitem[\protect\citeauthoryear{Takikawa, Evans, Tremblay, Müller, McGuire,
  Jacobson, and Fidler}{Takikawa et~al\mbox{.}}{2022}]%
        {vbnerf}
\bibfield{author}{\bibinfo{person}{Towaki Takikawa}, \bibinfo{person}{Alex
  Evans}, \bibinfo{person}{Jonathan Tremblay}, \bibinfo{person}{Thomas
  Müller}, \bibinfo{person}{Morgan McGuire}, \bibinfo{person}{Alec Jacobson},
  {and} \bibinfo{person}{Sanja Fidler}.} \bibinfo{year}{2022}\natexlab{}.
\newblock \showarticletitle{Variable Bitrate Neural Fields}. In
  \bibinfo{booktitle}{\emph{SIGGRAPH}}.
\newblock


\bibitem[\protect\citeauthoryear{Tretschk, Tewari, Golyanik, Zollh\"ofer,
  Lassner, and Theobalt}{Tretschk et~al\mbox{.}}{2021a}]%
        {Tretschk_2021_ICCV}
\bibfield{author}{\bibinfo{person}{Edgar Tretschk}, \bibinfo{person}{Ayush
  Tewari}, \bibinfo{person}{Vladislav Golyanik}, \bibinfo{person}{Michael
  Zollh\"ofer}, \bibinfo{person}{Christoph Lassner}, {and}
  \bibinfo{person}{Christian Theobalt}.} \bibinfo{year}{2021}\natexlab{a}.
\newblock \showarticletitle{Non-Rigid Neural Radiance Fields: Reconstruction
  and Novel View Synthesis of a Dynamic Scene From Monocular Video}. In
  \bibinfo{booktitle}{\emph{ICCV}}.
\newblock


\bibitem[\protect\citeauthoryear{Tretschk, Tewari, Golyanik, Zollh{\"o}fer,
  Lassner, and Theobalt}{Tretschk et~al\mbox{.}}{2021b}]%
        {tretschk2021non}
\bibfield{author}{\bibinfo{person}{Edgar Tretschk}, \bibinfo{person}{Ayush
  Tewari}, \bibinfo{person}{Vladislav Golyanik}, \bibinfo{person}{Michael
  Zollh{\"o}fer}, \bibinfo{person}{Christoph Lassner}, {and}
  \bibinfo{person}{Christian Theobalt}.} \bibinfo{year}{2021}\natexlab{b}.
\newblock \showarticletitle{Non-rigid neural radiance fields: Reconstruction
  and novel view synthesis of a dynamic scene from monocular video}. In
  \bibinfo{booktitle}{\emph{ICCV}}.
\newblock


\bibitem[\protect\citeauthoryear{Verbin, Hedman, Mildenhall, Zickler, Barron,
  and Srinivasan}{Verbin et~al\mbox{.}}{2022}]%
        {verbin2022refnerf}
\bibfield{author}{\bibinfo{person}{Dor Verbin}, \bibinfo{person}{Peter Hedman},
  \bibinfo{person}{Ben Mildenhall}, \bibinfo{person}{Todd Zickler},
  \bibinfo{person}{Jonathan~T. Barron}, {and} \bibinfo{person}{Pratul~P.
  Srinivasan}.} \bibinfo{year}{2022}\natexlab{}.
\newblock \showarticletitle{{Ref-NeRF}: Structured View-Dependent Appearance
  for Neural Radiance Fields}.
\newblock \bibinfo{journal}{\emph{CVPR}} (\bibinfo{year}{2022}).
\newblock


\bibitem[\protect\citeauthoryear{Wang, Zhang, Liu, Zhao, Zhang, Zhang, Wu, Xu,
  and Yu}{Wang et~al\mbox{.}}{2022}]%
        {wang2022fourier}
\bibfield{author}{\bibinfo{person}{Liao Wang}, \bibinfo{person}{Jiakai Zhang},
  \bibinfo{person}{Xinhang Liu}, \bibinfo{person}{Fuqiang Zhao},
  \bibinfo{person}{Yanshun Zhang}, \bibinfo{person}{Yingliang Zhang},
  \bibinfo{person}{Minye Wu}, \bibinfo{person}{Lan Xu}, {and}
  \bibinfo{person}{Jingyi Yu}.} \bibinfo{year}{2022}\natexlab{}.
\newblock \showarticletitle{Fourier PlenOctrees for Dynamic Radiance Field
  Rendering in Real-time}.
\newblock \bibinfo{journal}{\emph{arXiv:2202.08614}} (\bibinfo{year}{2022}).
\newblock


\bibitem[\protect\citeauthoryear{Wang, Wang, Genova, Srinivasan, Zhou, Barron,
  Martin-Brualla, Snavely, and Funkhouser}{Wang et~al\mbox{.}}{2021a}]%
        {wang2021ibrnet}
\bibfield{author}{\bibinfo{person}{Qianqian Wang}, \bibinfo{person}{Zhicheng
  Wang}, \bibinfo{person}{Kyle Genova}, \bibinfo{person}{Pratul~P Srinivasan},
  \bibinfo{person}{Howard Zhou}, \bibinfo{person}{Jonathan~T Barron},
  \bibinfo{person}{Ricardo Martin-Brualla}, \bibinfo{person}{Noah Snavely},
  {and} \bibinfo{person}{Thomas Funkhouser}.} \bibinfo{year}{2021}\natexlab{a}.
\newblock \showarticletitle{Ibrnet: Learning multi-view image-based rendering}.
  In \bibinfo{booktitle}{\emph{CVPR}}.
\newblock


\bibitem[\protect\citeauthoryear{Wang, Bovik, Sheikh, and Simoncelli}{Wang
  et~al\mbox{.}}{2004}]%
        {wang2004image}
\bibfield{author}{\bibinfo{person}{Zhou Wang}, \bibinfo{person}{Alan~C Bovik},
  \bibinfo{person}{Hamid~R Sheikh}, {and} \bibinfo{person}{Eero~P Simoncelli}.}
  \bibinfo{year}{2004}\natexlab{}.
\newblock \showarticletitle{Image quality assessment: from error visibility to
  structural similarity}.
\newblock \bibinfo{journal}{\emph{IEEE TIP}} (\bibinfo{year}{2004}).
\newblock


\bibitem[\protect\citeauthoryear{Wang, Simoncelli, and Bovik}{Wang
  et~al\mbox{.}}{2003}]%
        {wang2003multiscale}
\bibfield{author}{\bibinfo{person}{Zhou Wang}, \bibinfo{person}{Eero~P
  Simoncelli}, {and} \bibinfo{person}{Alan~C Bovik}.}
  \bibinfo{year}{2003}\natexlab{}.
\newblock \showarticletitle{Multiscale structural similarity for image quality
  assessment}. In \bibinfo{booktitle}{\emph{The Thrity-Seventh Asilomar
  Conference on Signals, Systems \& Computers, 2003}}.
\newblock


\bibitem[\protect\citeauthoryear{Wang, Wu, Xie, Chen, and Prisacariu}{Wang
  et~al\mbox{.}}{2021b}]%
        {wang2021nerf}
\bibfield{author}{\bibinfo{person}{Zirui Wang}, \bibinfo{person}{Shangzhe Wu},
  \bibinfo{person}{Weidi Xie}, \bibinfo{person}{Min Chen}, {and}
  \bibinfo{person}{Victor~Adrian Prisacariu}.}
  \bibinfo{year}{2021}\natexlab{b}.
\newblock \showarticletitle{NeRF--: Neural radiance fields without known camera
  parameters}.
\newblock \bibinfo{journal}{\emph{arXiv:2102.07064}} (\bibinfo{year}{2021}).
\newblock


\bibitem[\protect\citeauthoryear{Weng, Curless, Srinivasan, Barron, and
  Kemelmacher-Shlizerman}{Weng et~al\mbox{.}}{2022}]%
        {weng2022humannerf}
\bibfield{author}{\bibinfo{person}{Chung-Yi Weng}, \bibinfo{person}{Brian
  Curless}, \bibinfo{person}{Pratul~P Srinivasan}, \bibinfo{person}{Jonathan~T
  Barron}, {and} \bibinfo{person}{Ira Kemelmacher-Shlizerman}.}
  \bibinfo{year}{2022}\natexlab{}.
\newblock \showarticletitle{Humannerf: Free-viewpoint rendering of moving
  people from monocular video}. In \bibinfo{booktitle}{\emph{CVPR}}.
\newblock


\bibitem[\protect\citeauthoryear{Wizadwongsa, Phongthawee, Yenphraphai, and
  Suwajanakorn}{Wizadwongsa et~al\mbox{.}}{2021}]%
        {wizadwongsa2021nex}
\bibfield{author}{\bibinfo{person}{Suttisak Wizadwongsa},
  \bibinfo{person}{Pakkapon Phongthawee}, \bibinfo{person}{Jiraphon
  Yenphraphai}, {and} \bibinfo{person}{Supasorn Suwajanakorn}.}
  \bibinfo{year}{2021}\natexlab{}.
\newblock \showarticletitle{Nex: Real-time view synthesis with neural basis
  expansion}. In \bibinfo{booktitle}{\emph{CVPR}}.
\newblock


\bibitem[\protect\citeauthoryear{Xian, Huang, Kopf, and Kim}{Xian
  et~al\mbox{.}}{2021}]%
        {xian2020space}
\bibfield{author}{\bibinfo{person}{Wenqi Xian}, \bibinfo{person}{Jia-Bin
  Huang}, \bibinfo{person}{Johannes Kopf}, {and} \bibinfo{person}{Changil
  Kim}.} \bibinfo{year}{2021}\natexlab{}.
\newblock \showarticletitle{Space-time Neural Irradiance Fields for
  Free-Viewpoint Video}. In \bibinfo{booktitle}{\emph{CVPR}}.
\newblock


\bibitem[\protect\citeauthoryear{Xu, Alldieck, and Sminchisescu}{Xu
  et~al\mbox{.}}{2021}]%
        {xu2021h}
\bibfield{author}{\bibinfo{person}{Hongyi Xu}, \bibinfo{person}{Thiemo
  Alldieck}, {and} \bibinfo{person}{Cristian Sminchisescu}.}
  \bibinfo{year}{2021}\natexlab{}.
\newblock \showarticletitle{H-nerf: Neural radiance fields for rendering and
  temporal reconstruction of humans in motion}.
\newblock \bibinfo{journal}{\emph{NeurIPS}} (\bibinfo{year}{2021}).
\newblock


\bibitem[\protect\citeauthoryear{Yu, Fridovich-Keil, Tancik, Chen, Recht, and
  Kanazawa}{Yu et~al\mbox{.}}{2022}]%
        {yu_and_fridovichkeil2021plenoxels}
\bibfield{author}{\bibinfo{person}{Alex Yu}, \bibinfo{person}{Sara
  Fridovich-Keil}, \bibinfo{person}{Matthew Tancik}, \bibinfo{person}{Qinhong
  Chen}, \bibinfo{person}{Benjamin Recht}, {and} \bibinfo{person}{Angjoo
  Kanazawa}.} \bibinfo{year}{2022}\natexlab{}.
\newblock \showarticletitle{Plenoxels: Radiance Fields without Neural
  Networks}. In \bibinfo{booktitle}{\emph{CVPR}}.
\newblock


\bibitem[\protect\citeauthoryear{Yu, Li, Tancik, Li, Ng, and Kanazawa}{Yu
  et~al\mbox{.}}{2021a}]%
        {Yu_2021_ICCV}
\bibfield{author}{\bibinfo{person}{Alex Yu}, \bibinfo{person}{Ruilong Li},
  \bibinfo{person}{Matthew Tancik}, \bibinfo{person}{Hao Li},
  \bibinfo{person}{Ren Ng}, {and} \bibinfo{person}{Angjoo Kanazawa}.}
  \bibinfo{year}{2021}\natexlab{a}.
\newblock \showarticletitle{PlenOctrees for Real-Time Rendering of Neural
  Radiance Fields}. In \bibinfo{booktitle}{\emph{ICCV}}.
\newblock


\bibitem[\protect\citeauthoryear{Yu, Ye, Tancik, and Kanazawa}{Yu
  et~al\mbox{.}}{2021b}]%
        {yu2021pixelnerf}
\bibfield{author}{\bibinfo{person}{Alex Yu}, \bibinfo{person}{Vickie Ye},
  \bibinfo{person}{Matthew Tancik}, {and} \bibinfo{person}{Angjoo Kanazawa}.}
  \bibinfo{year}{2021}\natexlab{b}.
\newblock \showarticletitle{pixelNeRF: Neural radiance fields from one or few
  images}. In \bibinfo{booktitle}{\emph{CVPR}}.
\newblock


\bibitem[\protect\citeauthoryear{Zhang, Riegler, Snavely, and Koltun}{Zhang
  et~al\mbox{.}}{2020}]%
        {kaizhang2020}
\bibfield{author}{\bibinfo{person}{Kai Zhang}, \bibinfo{person}{Gernot
  Riegler}, \bibinfo{person}{Noah Snavely}, {and} \bibinfo{person}{Vladlen
  Koltun}.} \bibinfo{year}{2020}\natexlab{}.
\newblock \showarticletitle{NeRF++: Analyzing and Improving Neural Radiance
  Fields}.
\newblock \bibinfo{journal}{\emph{arXiv:2010.07492}} (\bibinfo{year}{2020}).
\newblock


\bibitem[\protect\citeauthoryear{Zhang, Isola, Efros, Shechtman, and
  Wang}{Zhang et~al\mbox{.}}{2018}]%
        {zhang2018unreasonable}
\bibfield{author}{\bibinfo{person}{Richard Zhang}, \bibinfo{person}{Phillip
  Isola}, \bibinfo{person}{Alexei~A Efros}, \bibinfo{person}{Eli Shechtman},
  {and} \bibinfo{person}{Oliver Wang}.} \bibinfo{year}{2018}\natexlab{}.
\newblock \showarticletitle{The unreasonable effectiveness of deep features as
  a perceptual metric}. In \bibinfo{booktitle}{\emph{CVPR}}.
\newblock


\end{thebibliography}

\appendix

\section{Appendix}
\subsection{More Ablation Studies}

\begin{figure}[t!]
    \centering
    \includegraphics[width=\linewidth]{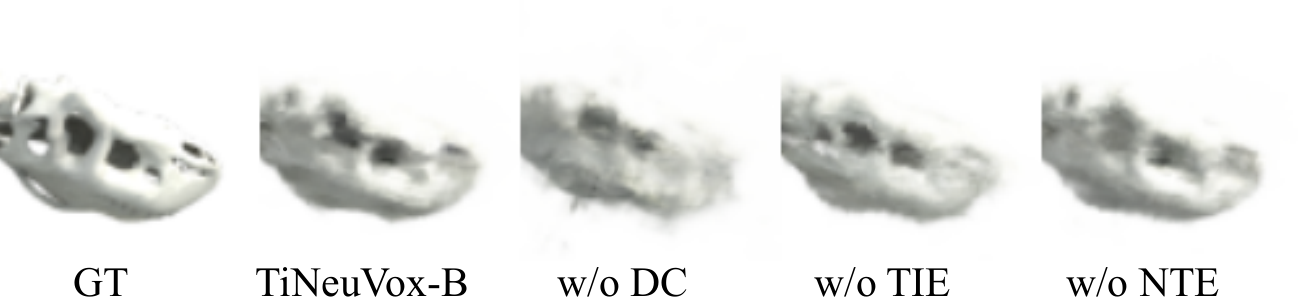}
    \caption{Qualitative comparisons by ablating three time-encoding components of \name. ``DC'' denotes deforming coordinates, ``TIE'' denotes temporal information enhancement, and ``NTE'' denotes neural time embeddings.}
    \label{fig: compts-imgs}
\end{figure}

\begin{figure}[t!]
    \centering
    \includegraphics[width=\linewidth]{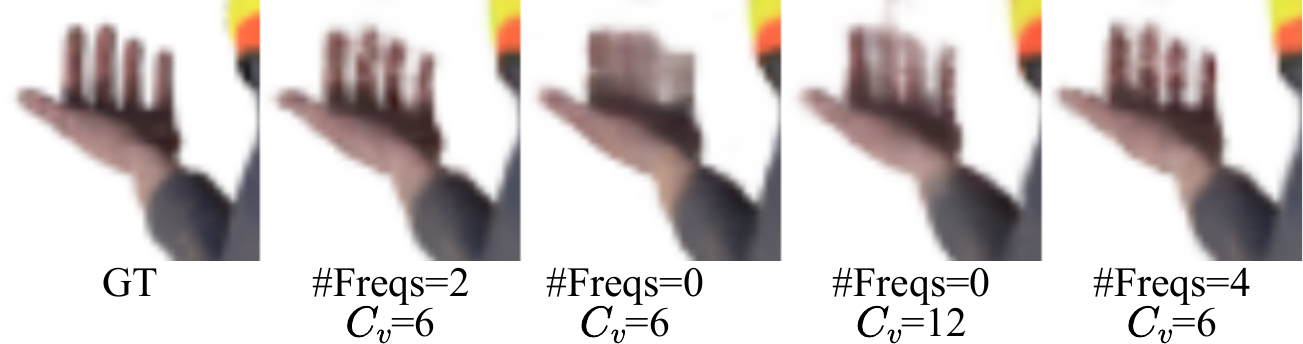}
    \caption{Qualitative studies about positional encoding on neural voxels.}
    \label{fig: pe-imgs}
\end{figure}

\begin{table}[t!]
\newcommand{\xmark}{\ding{55}}
\centering
\caption{Quantitative study about positional encoding on neural voxels with \name-B. $\#$Freqs. denotes the frequency number of positional encoding. $C_v$ denotes the channel number of neural voxels.}
\setlength{\tabcolsep}{4pt}
\label{tab: abla-pe}
\begin{tabular}{cc|cc|ccc}
\toprule
\textbf{\#Freqs.} & $\bm C_v$ & \textbf{Time} & \textbf{Storage} & \textbf{PSNR} $\uparrow$ & \textbf{SSIM} $\uparrow$ & \textbf{LPIPS} $\downarrow$\\
\midrule
2 & 6 & 28 mins & 48 MB & \textbf{32.668} & \textbf{0.971} & \textbf{0.041}\\
0 & 6 &  25 mins &48 MB  &32.255& 0.970& 0.051\\
0 & 12 &  28 mins & 94 MB& 32.440&0.971 & 0.045 \\
4 & 6 &  31 mins & 48 MB &32.111 & 0.970 & 0.047\\
\bottomrule
\end{tabular}
\end{table}

\paragraph{Positional Encoding on Neural Voxels}
We observe that applying positional encoding (PE) on neural voxels plays an important role in compressing voxel channel dimensions $C_v$. Removing PE on neural voxels results in dramatic performance degradation as in Tab.~\ref{tab: abla-pe} and details missing as in Fig.~\ref{fig: pe-imgs}. We then enlarge $C_v$ to $12$. Even though the storage cost increases to $94$ MB, the rendering performance is still worse than the smaller $C_v = 6$ with $2$-frequency PE. Above experiments demonstrate that PE on neural voxels has great power on modelling details with limited channel dimensions, which requires no additional parameters. We further experiment with a larger frequency number $4$ which leads to a worse result.

\begin{table}[t!]
\newcommand{\xmark}{\ding{55}}
\centering
\caption{Ablation study about hidden-layer widths $C_h$ on synthetic scenes.}
\begin{tabular}{c|c|ccc}
\toprule
\textbf{Width} & \textbf{Time}&\textbf{PSNR} $\uparrow$ & \textbf{SSIM} $\uparrow$ & \textbf{LPIPS} $\downarrow$ \\
\midrule
64 & 13 mins & 31.747 & 0.966 & 0.056 \\
128  &  19 mins& 32.040 & 0.968 & 0.048 \\
256 & 28 mins & \textbf{32.668} & \textbf{0.971} & \textbf{0.041}\\
\bottomrule
\end{tabular}
\label{tab: abla-width}
\end{table}

\paragraph{Hidden Layer Widths}
We study the widths of hidden-layer MLPs on \name-B and provide results in Tab.~\ref{tab: abla-width}. Larger-width MLPs bring better rendering performance. Though implicit representations have great power for modelling scenes, they accounts for a large portion of computation cost as they are required for every sample along rays and result in longer training time.

\begin{table}[t!]
\newcommand{\xmark}{\ding{55}}
\centering
\caption{Ablation study about feature dimensions $C_v$ of neural voxels.}
\begin{tabular}{c|cc|ccc}
\toprule
\textbf{$\bm C_v$} & \textbf{Time} & \textbf{Storage} & \textbf{PSNR} $\uparrow$ & \textbf{SSIM} $\uparrow$ & \textbf{LPIPS} $\downarrow$ \\
\midrule
4 & 25 mis& 32 MB & 32.318 & 0.970 & 0.048 \\
6 & 28 mins & 48 MB & 32.668 & 0.971 & 0.041\\
12 & 36 mins& 94 MB & 32.377 & 0.973 & 0.039 \\
\bottomrule
\end{tabular}
\label{tab: abla-voxel-width}
\end{table}

\paragraph{Neural Voxel Dimensions}
We study channel dimensions of neural voxels and show quantitative results in Tab.~\ref{tab: abla-voxel-width}. As in Row 2, we take \name-B as the default setting, \ie $C_v = 6$. Then in Row 1 and 3, we respectively decrease and increase the channel dimensions of neural voxels to $4$ and $12$. Performance gain can be obtained with more channels, while time and storage cost is slightly increased accordingly. Though the setting with $C_v = 12$ does not get a higher PSNR, it obtains better SSIM and LPIPS metrics.

\begin{figure}[t]
  \centering
  \includegraphics[width=\linewidth]{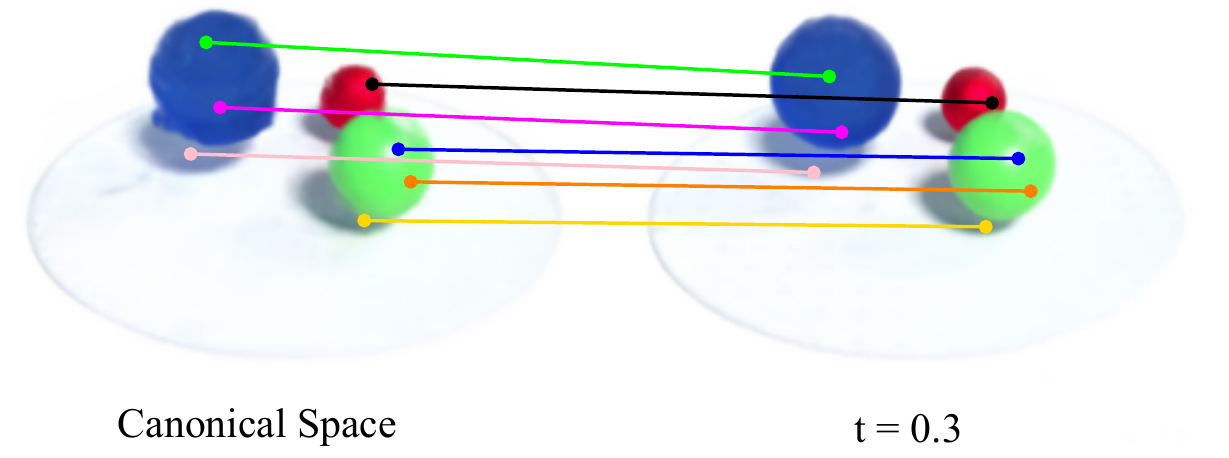}
  \caption{\rev{Visualization of Coordinate Deformation. Points in the canonical space (\emph{left}) are mapped into the $t=0.3$ space (\emph{right}), where $t$ denotes the time stamp. Note that as we do not explicitly indicate the canonical space (which we find will lead to better performance), the shown canonical space is not as clear as an specific-time one.}}
  \label{fig: vis-deform}
\end{figure}

\begin{table}[t!]
\newcommand{\xmark}{\ding{55}}
\centering
\caption{\rev{Comparisons between building separate pyramid voxel grids and multi-distance interpolating (MDI) one single voxel set.}}
\setlength{\tabcolsep}{4pt}
\label{tab: abla-pyramid}
\begin{tabular}{l|cc|ccc}
\toprule
\textbf{Method} & \textbf{Time} & \textbf{Storage} & \textbf{PSNR} $\uparrow$ & \textbf{SSIM} $\uparrow$ & \textbf{LPIPS} $\downarrow$\\
\midrule
Sep. Pyramid & 29 mins & 55 MB  & 32.69 & 0.97 & 0.04 \\
MDI (ours) &  28 mins & 48 MB & 32.67 & 0.97  & 0.04 \\
\bottomrule
\end{tabular}
\end{table}

\begin{figure}[t!]
  \centering
  \includegraphics[width=\linewidth]{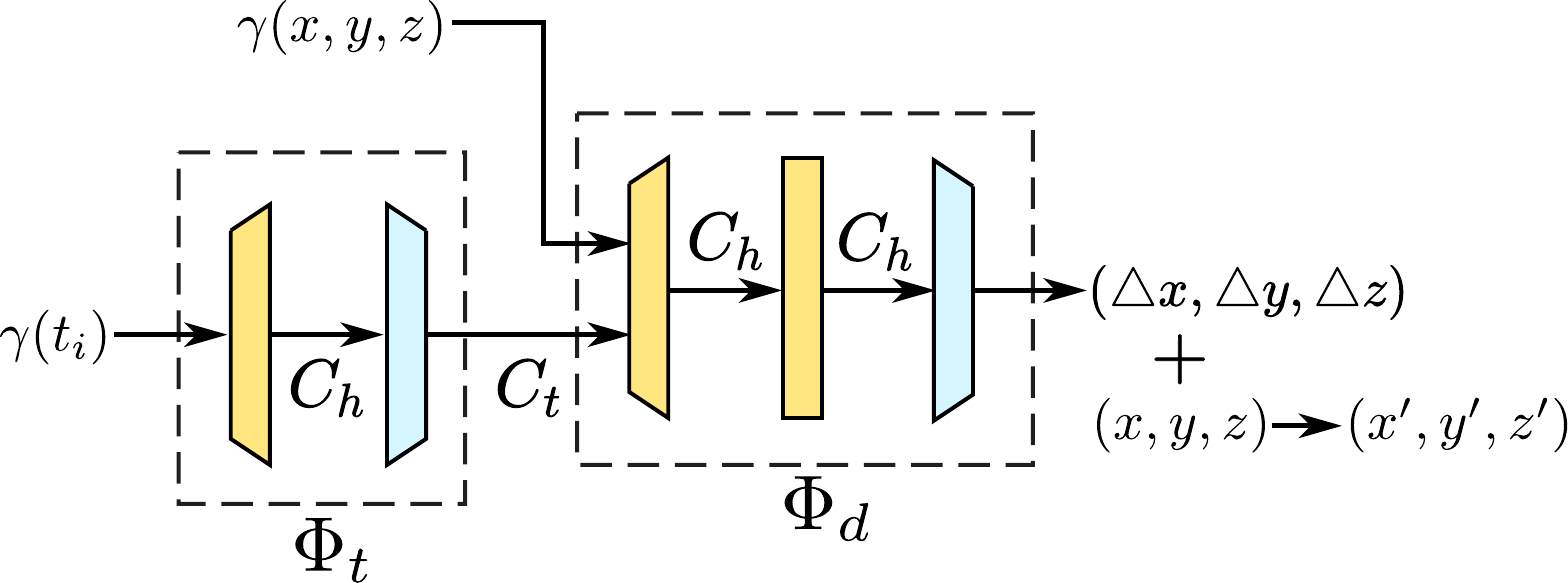}
  \caption{Architecture of the deformation network $\Phi_d$ along with the time-encoding network $\Phi_t$. $\gamma(\cdot)$ denotes the positional encoding. Yellow and blue boxes denote MLPs with and without ReLU activation. ``$C_h$'' denotes channel dimensions for hidden layers and ``$C_t$'' is the output channel dimensions of the time-encoding network $\Phi_t$.}
  \label{fig: deform}
\end{figure}

\begin{figure}[t!]
  \centering
  \includegraphics[width=\linewidth]{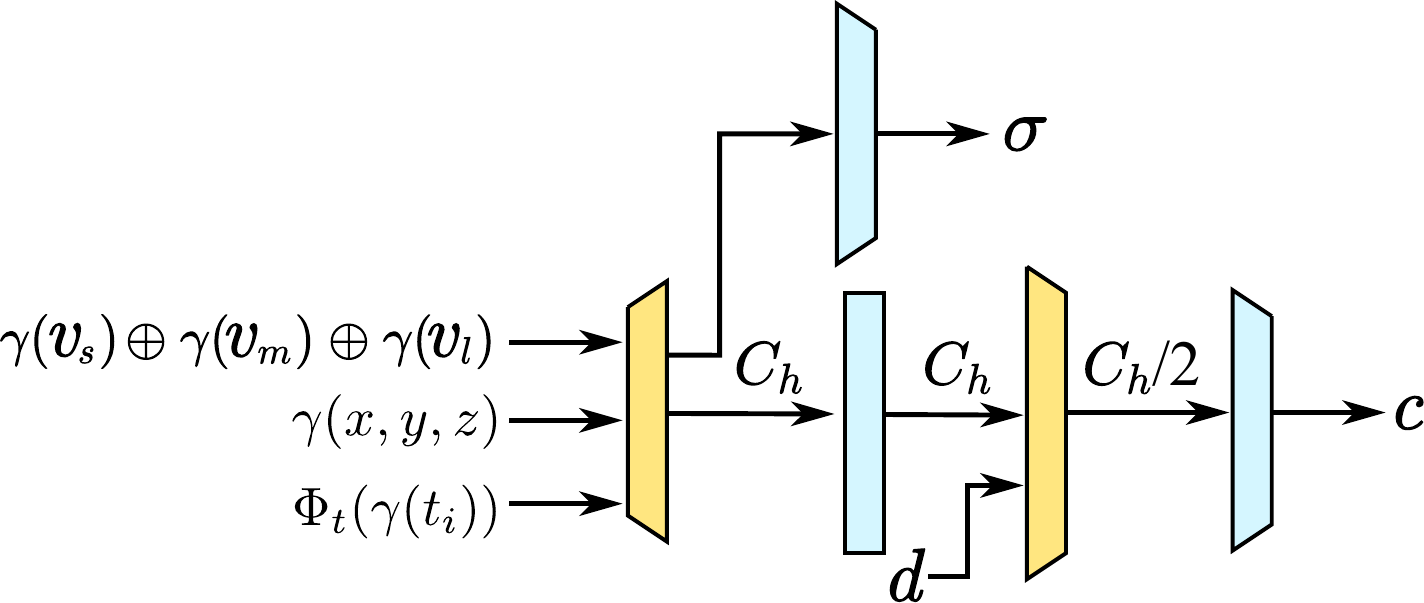}
  \caption{Architecture of the radiance network $\Phi_r$. ``$f_l$'', ``$f_m$'' and ``$f_s$'' denote voxel features interpolated from sampled grids with large, medium and small strides respectively. ``$\vd$'', ``$\vc$'' and ``$\sigma$'' denote view direction, color vectors and the density value.}
  \label{fig: radiance}
\end{figure}

\begin{table*}[t!]
\setlength{\tabcolsep}{3pt}
\centering
\caption{Per-scene quantitative comparisons on synthetic dynamic scenes.}
\label{table: synth-comp}
\begin{tabular}{lcccccccccccc}
\toprule
& \multicolumn{3}{c}{Hell Warrior} & \multicolumn{3}{c}{Mutant} & \multicolumn{3}{c}{Hook} & \multicolumn{3}{c}{Bouncing Balls} \\
\textbf{Method} 
 & PSNR$\uparrow$  & SSIM$\uparrow$ & LPIPS$\downarrow$
& PSNR$\uparrow$  & SSIM$\uparrow$ & LPIPS$\downarrow$
 & PSNR$\uparrow$  & SSIM$\uparrow$ & LPIPS$\downarrow$
 & PSNR$\uparrow$  & SSIM$\uparrow$ & LPIPS$\downarrow$\\
\cmidrule(lr){1-1}  \cmidrule(lr){2-4} \cmidrule(lr){5-7} \cmidrule(lr){8-10} \cmidrule(lr){11-13}
    NeRF~\cite{mildenhall2020nerf}
     & 13.52 & 0.81 & 0.25
     & 20.31 & 0.91 & 0.09
     & 16.65 & 0.84 & 0.19
    & 20.26 & 0.91 & 0.20 \\
    DirectVoxGo~\cite{sun2021direct}
    & 13.32 & 0.75 & 0.25
    &  19.45& 0.89 & 0.12
    & 16.16 & 0.80 &0.21 
    & 20.20& 0.87 & 0.22 \\
    Plenoxels~\cite{yu_and_fridovichkeil2021plenoxels}
    & 15.19 &0.78  &0.27 
    & 21.44 & 0.91 & 0.09
    & 17.90 & 0.81 &0.21 
    & 21.30& 0.89 & 0.18 \\
    T-NeRF~\cite{pumarola2021d}
    & 23.19 & 0.93 & 0.08
    & 30.56 & 0.96 & 0.04
    & 27.21 & 0.94 & 0.06
    & 37.81 & 0.98 & 0.12 \\
    D-NeRF~\cite{pumarola2021d}
     & 25.02 & 0.95 & \textbf{0.06}
     & 31.29 & 0.97 & \textbf{0.02}
    & 29.25 & 0.96 & 0.11
    & 38.93 & 0.98 & 0.10 \\
    \name-S (ours)
    & 27.00 &0.95  & 0.09 
    & 31.09 &0.96  &0.05 
    & 29.30 & 0.95 &0.07 
    & 39.05& 0.99 &0.06  \\
    \name-B (ours)
    &  \textbf{28.17}&\textbf{0.97 } &0.07 
    &  \textbf{33.61}&\textbf{0.98}  &0.03 
    &  \textbf{31.45}&\textbf{0.97 } &\textbf{0.05 }
    & \textbf{40.73}&\textbf{0.99}  &\textbf{0.04}  \\

\midrule
& \multicolumn{3}{c}{Lego} & \multicolumn{3}{c}{T-Rex} & \multicolumn{3}{c}{Stand Up} & \multicolumn{3}{c}{Jumping Jacks}  \\ \textbf{Method}
 & PSNR$\uparrow$  & SSIM$\uparrow$ & LPIPS$\downarrow$
 & PSNR$\uparrow$  & SSIM$\uparrow$ & LPIPS$\downarrow$
 & PSNR$\uparrow$  & SSIM$\uparrow$ & LPIPS$\downarrow$
& PSNR$\uparrow$  & SSIM$\uparrow$ & LPIPS$\downarrow$\\  
\cmidrule(lr){1-1}  \cmidrule(lr){2-4} \cmidrule(lr){5-7} \cmidrule(lr){8-10} \cmidrule(lr){11-13}
    NeRF 
    & 20.30 & 0.79 & 0.23
    & 24.49 & 0.93 & 0.13
    & 18.19 & 0.89 & 0.14
    & 18.28 & 0.88 & 0.23 \\
    DirectVoxGo~\cite{sun2021direct}
    & 21.13 &0.90  &0.10 
    & 23.27 &0.92  &0.09 
    & 17.58 &0.86  &0.16 
    & 17.80&0.84  &0.20  \\
    Plenoxels~\cite{yu_and_fridovichkeil2021plenoxels}
    & 21.97 & 0.90 &0.11 
    &  25.18& 0.93 & 0.08
    &  18.76& 0.87 & 0.15
    & 20.18&  0.86& 0.19 \\
    T-NeRF~\cite{pumarola2021d}
    & 23.82 & 0.90 & 0.15
    & 30.19 & 0.96 & 0.13
    & 31.24 & 0.97 & 0.02
    & 32.01 & 0.97 & 0.03 \\
    D-NeRF~\cite{pumarola2021d}
    & 21.64 & 0.83 & 0.16 
    & 31.75 & 0.97 & 0.03
    & 32.79 & 0.98 & 0.02
    & 32.80 & 0.98 & 0.03 \\
    \name-S (ours)
    & 24.35 &0.88  &0.13 
    & 29.95 &0.96  & 0.06
    & 32.89 &0.98  &0.03 
    & 32.33& 0.97 &0.04  \\
    \name-B (ours)
    &\textbf{25.02 } &\textbf{0.92}  &\textbf{0.07} 
    &\textbf{32.70} &\textbf{0.98}  &\textbf{0.03 }
    &\textbf{35.43}& \textbf{0.99} &\textbf{0.02 }
    & \textbf{34.23}& \textbf{0.98 }& \textbf{0.03} \\
\bottomrule
\end{tabular}
\end{table*}

\begin{table*}[thbp]
\setlength{\tabcolsep}{2pt}
\centering
\caption{Per-scene quantitative comparisons on real dynamic scenes.}
\label{tab: real-comp}
\begin{tabular}{l|c|cccccccc|cc}
\toprule 
\multirow{2}*{\textbf{Method}}
 & \multirow{2}*{\textbf{Time}}
& \multicolumn{2}{c}{Broom} & \multicolumn{2}{c}{3D Printer} & \multicolumn{2}{c}{Chicken} & \multicolumn{2}{c}{Peel Banana}  & \multicolumn{2}{|c}{\textbf{Mean}}\\ 
& & PSNR$\uparrow$  & MS-SSIM$\uparrow$ 
 & PSNR$\uparrow$  & MS-SSIM$\uparrow$ 
 & PSNR$\uparrow$  & MS-SSIM$\uparrow$ 
& PSNR$\uparrow$  & MS-SSIM$\uparrow$ 
& PSNR$\uparrow$  & MS-SSIM$\uparrow$ \\
\midrule
    NeRF~\cite{mildenhall2020nerf} & $\sim$ hours
    & 19.9&0.653 
    & 20.7&0.780 
    & 19.9&0.777
    & 20.0&0.769 
    & 20.1&0.745 \\
    NV~\cite{lombardi2019neural} & $\sim$ hours
    & 17.7&0.623 
    & 16.2& 0.665
    & 17.6& 0.615 
    & 15.9&0.380   
    & 16.9&0.571 \\
    NSFF~\cite{li2021nsff} & $\sim$ hours
    &26.1 &0.871 
    & 27.7& 0.947
    &26.9 & 0.944
    & 24.6& 0.902  
    & 26.3& 0.916\\
    Nerfies~\cite{park2021nerfies} & $\sim$ hours
    &19.2 &0.567 
    & 20.6&0.830 
    & 26.7& 0.943 
    & 22.4& 0.872 
    & 22.2& 0.803\\
    HyperNeRF~\cite{park2021hypernerf} & 32 hours$^\dagger$
    &19.3 &0.591 
    & 20.0& 0.821
    & 26.9& 0.948
    & 23.3&0.896  
    & 22.4& 0.814\\
    \midrule
    \name-S (ours)  & 10 mins
    & 21.9 &  0.707
    & 22.7 &0.836  
    & 27.0 & 0.929
    &22.1  &0.780
    & 23.4  & 0.813 \\
    
    \name-B (ours) &30 mins
    &21.5  &0.686   
    &22.8  &0.841  
    &28.3  &0.947 
    &24.4  &0.873  
    &24.3  &0.837  \\
\bottomrule
\end{tabular}
\begin{tablenotes}
\footnotesize
\item[] \textdagger \; Time cost of HyperNeRF~\cite{park2021hypernerf} is estimated according to descriptions in their paper but on TPUs.
\end{tablenotes}
\end{table*}

\rev{
\subsection{Visualization of Coordinate Deformation}
To analyze the quality of the deformation network, we visualize the coordinate deformation in Fig.~\ref{fig: vis-deform}. It can be observed that most points can be deformed to the same corresponding position but small drift may exist. This further reveals that the deformation network only predicts a coarse motion trajectory, where deviation it brings will be suppressed/eliminated by subsequent multi-distance interpolation and temporal information enhancement.
}

\begin{figure*}[thbp]
    \centering
    \includegraphics[width=0.9\linewidth]{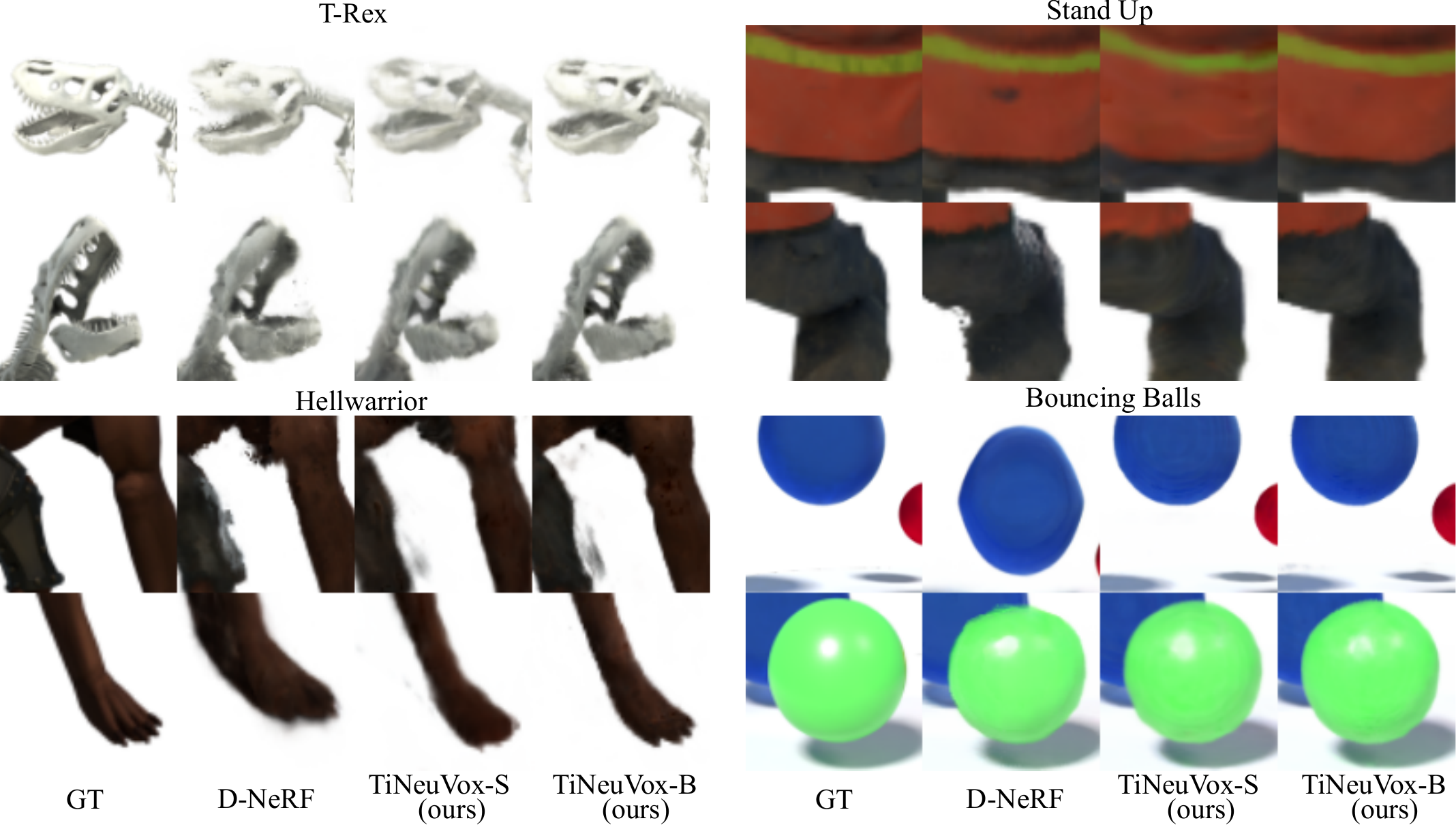}
    \caption{More qualitative comparisons between D-NeRF~\cite{pumarola2021d} and our \name on synthetic scenes. ``GT'' denotes groundtruth.}
    \label{fig: syth-imgs}
\end{figure*}

\revb{
\subsection{Comparisons with Separate Pyramid Voxels Grids}
We build a baseline of separate pyramid voxel grids to show efficiency of the proposed multi-distance interpolation (MDI) method. For this baseline, three separate voxel sets are constructed at different resolutions, \ie $160^3$, $80^3$, and $40^3$. Then three neural voxels, $\vv_l$, $\vv_m$ and $\vv_s$, are obtained by interpolating the three voxel sets respectively and concatenated to be fed into the radiance network. In \name, only one single set of voxels at the $160^3$ resolution is built. Three neural voxels are interpolated from the same voxel set with different sample distances. As shown in Tab.~\ref{tab: abla-pyramid}, compared with building separate pyramid voxel grids, MDI with one single set of voxels takes 13\% smaller storage cost with similar time but produces a same rendering quality.
}

\rev{
\subsection{Details of Additional Losses}
We follow the proposals in DirectVoxGo~\cite{sun2021direct} and adopt two additional losses for better training. As in Eq.~\ref{eq: sup_all}, one loss supervises all samples with the target color for stabilizing the optimization process to mitigate local minima, in particular during the initial training phase. Meanwhile, a small loss weight $10^{-2}$ is used, thus avoiding all samples producing the same color.
\begin{equation}
\label{eq: sup_all}
\mathcal{L}_\text{all\_pts} = \sum_{i=1}^N T_i(1 - \text{exp}(-\sigma_i \delta_i))\lVert\rvc_i-C(\rvr)\rVert^2_2.
\end{equation}
As shown in Eq.~\ref{eq: bg_loss}, the second one is a background entropy-loss which facilitates to better distinguish fore- and background areas, thus encouraging to focus on either region.
\begin{equation}
\label{eq: bg_loss}
\mathcal{L}_\text{bg} = -T_{N+1}\text{log}(T_{N+1}) - (1-T_{N+1})log(1-T_{N+1}),
\end{equation}
where $T_{N+1}$ denotes the rendered background probability. Our results show that the combination of these two losses leads to an 1.2-PSNR improvement.
}

\subsection{Neural Architectures}
We show architectures of the deformation network $\Phi_d$ along with the time-encoding network $\Phi_t$ in Fig.~\ref{fig: deform}, and the radiance network $\Phi_r$ in Fig.~\ref{fig: radiance}.

\subsection{Evaluation Results}
We show per-scene quantitative results for synthetic ones in Tab.~\ref{table: synth-comp} and real ones in Tab.~\ref{tab: real-comp}. Additional qualitative comparisons on synthetic scenes are provided in Fig.~\ref{fig: syth-imgs}. It is highly recommended to refer to the supplementary video for more detailed presentations.

\end{document}